\documentclass{article}

\usepackage{preamble}
\begin{document}

\maketitle
\begin{abstract}
The incorporation of appropriate inductive bias plays a critical role in learning dynamics from data. A growing body of work has been exploring ways to enforce energy conservation in the learned dynamics by encoding Lagrangian or Hamiltonian dynamics into the neural network architecture. These existing approaches are based on differential equations, which do not allow discontinuity in the states and thereby limit the class of systems one can learn. However, in reality, most physical systems, such as legged robots and robotic manipulators, involve contacts and collisions, which introduce discontinuities in the states. In this paper, we introduce a differentiable contact model, which can capture contact mechanics: frictionless/frictional, as well as elastic/inelastic. This model can also accommodate inequality constraints, such as limits on the joint angles. The proposed contact model extends the scope of Lagrangian and Hamiltonian neural networks by allowing simultaneous learning of contact and system properties. We demonstrate this framework on a series of challenging 2D and 3D physical systems with different coefficients of restitution and friction. The learned dynamics can be used as a differentiable physics simulator for downstream gradient-based optimization tasks, such as planning and control. \footnote{Code available at \url{https://github.com/Physics-aware-AI/DiffCoSim}.} \footnote{Video available at \url{https://www.youtube.com/watch?v=DdJ7RLmG0kg}.}
%Thus, by enabling simultaneous learning of contact properties, as well as system properties, the proposed contact model extends the scope of Hamiltonian and Lagrangian neural networks. 
%, both frictionless and frictional, as well as both elastic and inelastic.
%
% (e.g., coefficients of restitution and friction)
%Learning motion of physical systems requires appropriate inductive bias to learn the underlying dynamics and contact. Recent works explore the learning of dynamics by enforcing the prior of Lagrangian or Hamiltonian dynamics. These approaches assume the trajectories of the physical systems to be smooth, which limits the class of systems we can learn. Real systems such as legged robots and robot manipulators involve contacts and collisions which makes trajectories non-smooth. In this work, we introduce a differentiable contact model, which can capture frictionless and frictional contacts with \dz{various} elasticity as well as joint limits. The proposed contact model extends Lagrangian and Hamiltonian neural networks so that contact properties, i.e., coefficients of restitution and friction, as well as system properties can be learned simultaneously. We demonstrate this framework on a series of challenging 2D and 3D physical systems.
\end{abstract}

\section{Introduction}
% \dz{This paragraph may need to be rewritten. Motivate Contact and collision.}
%
%In a large class of real-world problems, the underlying physical systems evolve in a piecewise-continuous manner. For example, we can play tennis since tennis balls collide with the ground and rackets with high elasticity, but follow smooth trajectories governed by the laws of physics in between those collisions. Our ability to walk and run depends heavily on the frictional contacts between our shoes and the ground; icy roads are slippery since the contact surface is almost frictionless. Robotics systems, such as legged robots and manipulators, also rely on contacts to perform tasks. These scenarios highlight the relevance and importance of contacts and collisions, which exist everywhere, albeit with different properties.
%
A large class of real-world physical systems evolves in a piecewise-continuous manner. For example, while playing tennis, tennis balls collide with the ground and the rackets with high elasticity but follow smooth trajectories governed in between those collisions. The ability to walk/run depends heavily on the contacts between the legs and the ground. Unfavorable contact properties can significantly hinder this ability; for example, lack of friction makes it very difficult to walk on icy roads. Robotic manipulators and grippers also rely on contacts and collisions to perform their assigned tasks. These examples highlight the importance of contacts and collisions, which can be found everywhere.

Encoding energy conservation into the computation graph of a neural network constitutes an effective way to improve its data-efficiency and generalization performance in inferring the dynamics of a physical system from its trajectory data \citep{zhong2020benchmarking}. However, as these energy-conserving models assume the system trajectories to be smooth and governed by ordinary differential equations (ODE), they cannot model dynamics with contacts and collisions. On the other hand, another line of work, for example, interaction network \cite{battaglia2016interaction}, neural physics engine \cite{chang2016compositional}, and iterative neural projection \cite{yang2020learning} can model collisions and contacts and learn the associated properties. However, they are not ODE-based and hence cannot infer the continuous dynamics governing the smooth portion of the trajectories. A more recent work \cite{pmlr-v130-hochlehnert21a} has encoded a discrete form of the Euler-Lagrange equation while learning properties of frictionless 2D contacts. Learning of properties associated with frictional contacts and 3D contacts still remains a relatively underexplored topic in the literature.

% have been tested on NPE, IN, and INP
% \emph{The restriction on contact free systems makes it impossible to apply these energy-conserving approaches to physical systems with contacts}. 
%
% Another line of work, such as Neural Physics Engine (NPE) \cite{chang2016compositional} and Interaction Networks (IN) \cite{battaglia2016interaction}, use relational priors to learn motions of physical systems. These models are not ODE-based and can handle collisions and contacts. A recent work by \citet{yang2020learning} directly learns physical constraints and contact with an iterative neural projection (INP) algorithm. However, only frictionless 2D contacts have been tested on NPE, IN and INP. The learning of frictional contacts as well as 3D contact is underexplored in the literature. 

In this work, we introduce a contact model that can handle frictional contacts both with or without elasticity as well as enforces energy-conservation during the smooth portions of the trajectories. The scope of energy-conserving neural networks are extended by the contact model.
% This, by incorporating contacts and collisions into the energy-conserving neural networks, extends their scope. 
The contact model solves convex optimization problems to calculate the jump in velocity during contact. In order to use this contact model in deep learning tasks, we build upon the recent progress on differentiating through convex optimization problems \cite{agrawal2019differentiable}. We demonstrate the performance of the differentiable contact model in learning coefficients of restitution and friction associated with a variety of 2D and 3D contacts. In addition, we also demonstrate the framework as a differentiable physics simulator and test it in downstream planning tasks.

%By introducing a contact model, which can handle frictional contacts both with or without elasticity, this work incorporates contacts and collisions into the energy-conserving neural networks and extends their scope. The proposed contact model solves convex optimization problems to calculate the jump in velocity during contact. To make our contact model differentiable, we build upon the recent progress on differentiating through convex optimization problems \cite{agrawal2019differentiable}. We demonstrate the performance of our contact model in learning coefficients of restitution and friction associated with a variety of 2D and 3D contacts. In addition, we also demonstrate the usefulness of this differentiable contact model along with differentiable dynamics in downstream planning tasks.
% \\ \textbf{\cite{liang2020differentiable}}
% \dz{need to be clear that we are not learning contacts, we still need some information from the collision detector.} 

\subsection{Related Work}
\textbf{Lagrangian/Hamiltonian-inspired Neural Networks:}
In the last few years, an increasing volume of work has proposed neural network models to learn the underlying dynamics from data while enforcing energy conservation. This line of works leverage Lagrangian dynamics \citep{lutter2018deep, lutter2019deep, roehrl2020modeling, cranmer2020lagrangian, zhong2020unsupervised, allen2020lagnetvip} or Hamiltonian dynamics \citep{greydanus2019hamiltonian, Zhong2020Symplectic, zhong2020dissipative, Chen2020Symplectic} to incorporate the physics prior of energy conservation into deep learning. Recently, \citet{finzi2020simplifying} show that using Cartesian coordinates and enforcing explicit constraints improve learning in both Lagrangian and Hamiltonian settings. To learn the underlying dynamics governed by an ODE, many of these prior works have used Neural ODE \citep{chen2018neural} which can learn an ODE from observed trajectories. However, real systems often exhibit non-smooth trajectories caused by sudden/abrupt changes in the velocity due to contacts and collisions. Although Neural ODE based approaches have recently been extended for learning dynamics with jump discontinuities \citep{NEURIPS2019_59b1deff, gwak2020neural, herrera2021neural}, they cannot accommodate the physical constraints (e.g., maximum dissipation principle, non-negative normal force) associated with contacts. Among the energy-conserving neural networks, only \cite{Chen2020Symplectic} attempted to address contacts and collisions; to capture the elastic collision of a billiard ball, it manually reverses the momentum of the ball orthogonal to the contact surface. However, this specialized technique cannot be applied to frictional or inelastic contacts, or objects that can rotate.
\textbf{Contact Model:}
Our contact model shares similarity with the contact model of MuJoCo \cite{todorov2011convex,todorov2014convex,todorov2012mujoco}. However, there are three differences: (1) our contact model handles elastic contacts while MuJoCo only focuses on inelastic contacts; (2) MuJoCo solves the convex optimization problem with a generalization of the Projected Gauss-Seidel method while we leverage the open-source \verb!scs! solver \citep{o2016conic} to solve the optimization problem; (3) the dynamics in MuJoCo is described using generalized coordinates, while we use Cartesian coordinates, since it has been shown in previous work \cite{finzi2020simplifying}, the use of Cartesian coordinates improves the learning of system properties. Another category of contact models solve contact impulses by solving a linear complementarity problem (LCP) \cite{anitescu1997formulating}. Recently, a number of works \cite{de2018end, degrave2019differentiable, liang2020differentiable, pmlr-v120-song20a, song2020learning, werling2021fast} has proposed differentiable LCPs for downstream planning and control tasks. However, their performance on learning the contact properties has yet to be tested. 
%\dz{How well these differentiable LCPs are able to learn contact properties has yet to be tested.}
%
%

%
\textbf{Differentiable Simulation:}
The recent past has also witnessed a growing interest in differentiable physics simulation that can be used in many downstream tasks (e.g., parameter estimation, planning, and control) \citep{xu2019densephysnet, sanchez2020learning, Hu2020DiffTaichi, 10.1145/3414685.3417766, 10.1111/cgf.14104, NeuralSim_icra21}. \citet{jiang2018dataaugmented} use an LCP formulation to learn contact impulses for perfectly inelastic contacts. DiffTaichi \cite{Hu2020DiffTaichi} focuses on material point method and only provides intuitively simple contact mechanisms. The support for partially elastic frictional contacts is yet to be provided. \citet{10.1145/3414685.3417766} differentiates through the dynamics solver analytically.
% {\color{blue}\citet{10.1111/cgf.14104} develop a contact model using a variational formulation based on generalized velocities and soft constraints, and analyzes the impact of various system parameters on the accuracy of contact simulation.}
\citet{10.1111/cgf.14104} develop a compliant contact model and formulate an implicit time-stepping scheme for integrating the dynamics. Incremental Potential Contact (IPC) \citep{li2020incremental} uses a custom implicitly time-stepped solver to solve nonlinear intersection-free and inversion-free elastodynamics. 
However, it is not clear if unknown dynamics and contact properties can be learned using the implicit time-stepping scheme proposed in \cite{10.1111/cgf.14104, li2020incremental}. NeuralSim \cite{NeuralSim_icra21} formulates a nonlinear complementarity problem to learn contact impulses and then solves it using Projected Gauss-Siedel. GradSim \citep{murthy2021gradsim} uses a relaxed Coulomb model to learn contacts from video sequences. 
\citet{le2021differentiable} propose a differentiable physics simulator that can handle conic friction and elasticity. They demonstrate its ability for system identification on a simple 2D system. 
\citet{chen2021learning} propose neural event functions to handle instantaneous changes in a continuous system and test it on frictionless bouncing balls.
However, these prior differentiable simulation models do not focus on the energy aspect of the system and their performance on the prediction of system energy (of learned dynamics) has not been evaluated. 

\subsection{Contribution}
The main contribution of this work is three-fold. First, by introducing a differentiable contact model, we extend the scope of Lagrangian/Hamiltonian-inspired deep learning methods from collision-free systems to more realistic systems with contact and collisions. Second, we demonstrate the simultaneous learning of system and contact properties in a variety of physical systems by integrating the contact model with Constrained Lagrangian/Hamiltonian neural networks (CLNN/CHNN). We show that the learned contact properties, i.e., coefficients of restitution and friction, are interpretable and match the ground truth with high accuracy. Finally, the learned contact model with CLNN/CHNN can be used to solve downstream gradient-based optimization tasks.
%applicability
% First, we propose a differentiable contact model, which can capture frictionless/frictional, as well as elastic/inelastic contacts and collisions. In addition, it can also enforce limit-constraints on the state variables.
%
%

\section{Preliminaries}
\subsection{Rigid body dynamics without contacts}
%Consider a rigid body system with holonomic constraints (equality constraints, see Appendix~\ref{sec:equ-constr} for details) but free of collision and contact. The configuration of the system at time $t$ can be described by a set of coordinates $\mathbf{x}(t) = (x_1(t), x_2(t), ..., x_D(t))$. Then the time evolution of the rigid body system can be expressed as the following second-order ordinary differential equation (ODE),
Consider a rigid body system whose configuration at time $t$ is described by a set of coordinates $\mathbf{x}(t) := (x_1(t), x_2(t), ..., x_D(t))$. Then the time evolution of this system can be expressed as the following second-order ODE,
\begin{equation}
    \label{eqn:2nd-order-dyna}
    \ddot{\mathbf{x}} = \mathbf{h}(\mathbf{x}, \dot{\mathbf{x}}; \mathbf{p}_s),
\end{equation}
where $\mathbf{p}_s$ denote system properties, which may include inertia of objects $\mathbf{M}(\mathbf{x})$ and potential energy $\mathbf{V}(\mathbf{x})$. The vector-valued function $\mathbf{h}$ can be derived from the laws of physics, e.g, Lagrangian/Hamiltonian dynamics. By introducing $\mathbf{v} := \dot{\mathbf{x}}$, Eqn. \eqref{eqn:2nd-order-dyna} can be written as the following first-order ODE
\begin{equation}
    %\dot{\textbf{z}} = \Psi(\textbf{z}) \Rightarrow
    \label{eqn:1nd-order-dyna}
    \begin{pmatrix}
        \dot{\mathbf{x}} \\
        \dot{\mathbf{v}}
    \end{pmatrix} = 
    \begin{pmatrix}
        \mathbf{v} \\
        \mathbf{h}(\mathbf{x}, \mathbf{v}; \mathbf{p}_s)
    \end{pmatrix}
    = \mathbf{g}(\mathbf{x}, \mathbf{v}; \mathbf{p}_s).
\end{equation}
% The vector field $\mathbf{g}$ can be derived from the law of physics and it depends on system properties $\mathbf{p}_s$, which may include inertia of objects $\mathbf{M}(\mathbf{x})$, potential field $\mathbf{V}(\mathbf{x})$ and \dz{equality constraints $\Phi (\mathbf{x})$}. 
%
There are two popular choices for the coordinates $\mathbf{x}$ -- the generalized coordinates and the Cartesian coordinates. The generalized coordinates are usually chosen as a set of independent coordinates which implicitly enforces holonomic constraints (equality constraints, see Appendix~B for details) in the system. The Cartesian coordinates are in general not independent of each other, so that the holonomic constraints in the system must be enforced explicitly in the dynamics \eqref{eqn:1nd-order-dyna}. Although $\mathbf{g}$ is usually derived with generalized coordinates, this work uses Cartesian coordinates and explicit constraints \cite{finzi2020simplifying} to demonstrate the results. The proposed contact model is independent of the choice of coordinates and $\mathbf{g}$. We provide the expression of $\mathbf{g}$ used in this work in Appendix~B.
%The Cartesian coordinates are in general not independent of each other, so that the holonomic constraints in the system must be enforced explicitly in the dynamics \eqref{eqn:1nd-order-dyna}.
% Although $\mathbf{g}$ is usually derived from Lagrangian mechanics with generalized coordinates, here we demonstrate our results based on the Lagrangian and Hamiltonian formalism with Cartesian coordinates.
%In practice, the actual form of $\mathbf{g}$ is usually derived from Lagrangian mechanics. Although in this work, we demonstrate our results based on the Lagrangian and Hamiltonian formalism with Cartesian coordinates, the proposed contact model is independent of the set of coordinates and form of $\mathbf{g}$. 
%
% The actual form of $\mathbf{f}$ derived by Lagrangian and Hamiltonian dynamics with both choices as well as the learning problem has been discussed in \cite{finzi2020simplifying, zhong2020benchmarking}. Our proposed differentiable contact impulse solver in this work is independent of the form of dynamics \eqref{eqn:2nd-order-dyna}, although we demonstrate it along with the constrained Lagrangian dynamics on various systems. 
%
\subsection{Rigid body dynamics with contacts}
% \begin{algorithm}[tb]
%   \caption{Simulation with Contact}
%   \label{alg:sim-w-c}
% \begin{algorithmic}
%   \STATE {\bfseries Input:} Initial condition $(\mathbf{x}_0, \mathbf{v}_0)$, dynamics $\mathbf{f}(\mathbf{x},\mathbf{v})$, collision detector and contact impulse solver
%   \STATE Initialize $(\mathbf{x},\mathbf{v}) = (\mathbf{x}_0, \mathbf{v}_0)$, initialize trajectories $\mathcal{T}$
%   \FOR{$t=1$ {\bfseries to} $T$}
%   \STATE $(\mathbf{x}_t, \mathbf{v}_t) \leftarrow \verb!ODESolve!(\mathbf{f}, (\mathbf{x}, \mathbf{v}), t-1, t)$
%   \STATE $isCld \leftarrow \verb!CollisionDectector!(\mathbf{x}_t)$
%   \IF{$isCld$}
%   \STATE $\Delta \mathbf{v} \leftarrow \verb!ImpulseSolver!(\mathbf{x}_t, \mathbf{v}_t)$
%   \STATE $\mathbf{v}_t \leftarrow \mathbf{v}_t + \Delta \mathbf{v}$
%   \ENDIF
%   \STATE$\mathcal{T}.\verb!append!((\mathbf{x}_t, \mathbf{v}_t))$
%   \STATE $(\mathbf{x}, \mathbf{v}) \leftarrow (\mathbf{x}_t, \mathbf{v}_t)$
%   \ENDFOR
%   \STATE {\bfseries Output:} Trajectories $\mathcal{T}$
% \end{algorithmic}
% \end{algorithm}
% \begin{wrapfigure}[17]{L}{0.7\textwidth}
\begin{wrapfigure}[19]{r}{0.53\textwidth}
\begin{minipage}[t]{0.53\textwidth}
    \vspace{-5pt}
    \begin{algorithm}[H]
    \caption{Rigid Body Dynamics with Contact}
    \label{alg:sim-w-c}
    \SetEndCharOfAlgoLine{}
    \SetKwComment{Comment}{// }{}
    \SetKwInOut{Input}{Input}
    \Input{\\\hspace{-3.6em}\small
    \begin{tabular}[t]{l @{\hspace{.5em}} l}% 
    $t_0, t_1, ..., t_N$ & Sequence of time points  \\
    $(\mathbf{x}_0, \mathbf{v}_0)$ & Initial condition at $t_0$\\
    $\mathbf{p}_s = (\mathbf{M}(\mathbf{x}), V(\mathbf{x}))$ & System properties \\
    $\mathbf{p}_c = (\pmb{\mu}, \mathbf{e}_P)$ & Contact properties \\
    $\mathbf{g}(\mathbf{x},\mathbf{v} ; \mathbf{p}_s)$ & First-order system dynamics
    \end{tabular}\hspace{-0.5em}% 
    }
    Initialize output trajectories $\mathcal{T}=\{(\mathbf{x}_0, \mathbf{v}_0)\}$. \;
    \For{$i = 0 \to N-1$}{
    $(\mathbf{x}_{i+1}, \mathbf{v}_{i+1}) \leftarrow \texttt{ODESolve}(\mathbf{g}, (\mathbf{x}_i, \mathbf{v}_i), t_i, t_{i+1})$ \;
    Get active contacts $\mathbf{c}_a$ (collision detection) \;
    % $\mathbf{c}_a \leftarrow \texttt{CollisionDectection}(\mathbf{x}_{i+1})$ \;
      \uIf{exist active contacts}{
      $\Delta \mathbf{v} \leftarrow \texttt{ContactModel}(\mathbf{x}_{i\!+\!1}, \mathbf{v}_{i\!+\!1},\! \mathbf{c}_a,\! \mathbf{p}_s,\! \mathbf{p}_c)$\;
      $\mathbf{v}_{i+1} \leftarrow \mathbf{v}_{i+1} + \Delta \mathbf{v}$ \;
      }
    $\mathcal{T}\leftarrow \mathcal{T} \cup \{(\mathbf{x}_{i+1}, \mathbf{v}_{i+1})\}$ \;
    % $(\mathbf{x}, \mathbf{v}) \leftarrow (\mathbf{x}_t, \mathbf{v}_t)$ \;
    }
    % Output: Trajectories $\mathcal{T}$ \;
    \end{algorithm}
\end{minipage}
\end{wrapfigure} \normalsize
In robotics tasks, the above assumption of no collision and contact no longer holds. For example, legged robots move around through repeated collisions/contacts between the robot legs and the ground, and robot arms grasp objects by making frictional contact with them. The difficulty of modeling these phenomena is that they essentially make the dynamics discontinuous. For example, when a ball hits the ground, its velocity changes from pointing downward to pointing upward in an infinitesimally small period of time, which can be modeled as an instantaneous change in velocity $\Delta v$. In general, contacts, collisions, and joint limits can all be modeled in this way. Algorithm \ref{alg:sim-w-c} shows the general procedure of modeling rigid body with contacts, where a jump in velocity is calculated by a contact model whenever there exist active contacts.
%by colliding the legs with the ground 
%contact modelling 
%contacts 
%a short amount 

From a simulation perspective, with known system properties, contact properties (coefficients of friction and restitution), and vector field $\mathbf{g}$, the trajectory of the system can be simulated by Algorithm~\ref{alg:sim-w-c}. From a learning perspective, we frame the problem as learning unknown system and contact properties from a given set of trajectories given a model prior of vector field $\mathbf{g}$. In this case, we can parametrize the unknown system and contact properties $(\mathbf{p}_s, \mathbf{p}_c)$ by neural networks and learnable parameters, predict trajectories by Algorithm \ref{alg:sim-w-c} and minimize the difference between the predicted and actual trajectories by backpropagation. This training scheme requires all operations in the forward pass (Algorithm \ref{alg:sim-w-c}) to be differentiable. There are two key parts in the forward pass -- the ODE solver module and the contact model. Operations in the ODE solver are in general differentiable, and Neural ODE \cite{chen2018neural, NEURIPS2020_dissectNeuralODE} provides a framework of backpropagating through ODE solvers with constant memory cost. In this work, we provide a differentiable contact model so that we can extend these previous works to learn rigid body dynamics with contacts. 
\section{A differentiable contact model}
\label{sec:DiffContactModel}
% We consider rigid body systems with equality constraints and contacts. 
In this section, we introduce a differentiable contact model for learning rigid body dynamics with contacts. The proposed contact model solves post-contact velocities by solving contact impulses in two phases \cite{poisson1817mechanics} -- a \emph{compression phase}, starting from the first contact of objects till the maximum compression, and a \emph{restitution phase}, starting right after the compression phase till the separation of objects. We start by presenting the constraints imposed by frictional contacts. 
\subsection{Contact constraints}
This work focuses on two types of contact -- frictional contact and limit constraint. 
For any conceptual frictional contact $i$ in a 3D contact space, the contact impulse $\mathbf{f}_i \in \mathbb{R}^3$ must lie in the friction cone, 
\begin{align}
    \label{eqn:2nd-cone-constraint}
    \mu_i f_{i,n} \geq \sqrt{f_{i,t_1}^2 + f_{i,t_2}^2}, \quad \forall i,
\end{align}
where $\mu_i\geq 0$ is the coefficient of friction for conceptual contact $i$. In addition, the normal impulses must be non-negative, since objects can only push but not pull others:
\begin{align}
    \label{eqn:non-neg-constraint}
    f_{i, n} \geq 0, \quad \forall i.
\end{align}
For any limit constraint such as limit in joint angle or distance, the contact space is essentially one dimensional and the constraints on contact impulses are only $f_{i, n} \geq 0$. This is mathematically equivalent to setting up a 3D contact space like that in frictional contact and letting $\mu_i=0$ in Eqn.~\eqref{eqn:2nd-cone-constraint}. 
\subsection{Contact model in compression phase}
The idea behind solving contact impulses during the compression phase is the maximum dissipation principle \cite{stewart2000rigid}, which states that the compression impulses should maximize the rate of energy dissipation. Equivalently, the compression impulses are those that minimizes the kinetic energy at the end of the compression phase. This can be described by an optimization problem \cite{yang2020learning, stewart2000rigid, todorov2011convex, todorov2014convex}. We choose the following form, which is similar to the one used in Mujoco \cite{todorov2011convex, todorov2014convex,todorov2012mujoco},
\begin{align}
    \label{eqn:comp-cvx-prob}
    &\underset{\mathbf{f}_C^c}{\textrm{Minimize }} \frac{1}{2} (\mathbf{f}_C^{c})^T \mathbf{A} \mathbf{f}_C^{c} + (\mathbf{f}_C^{c})^T \mathbf{v}_C^{c-} \\
    &\textrm{subject to } \eqref{eqn:2nd-cone-constraint}, \eqref{eqn:non-neg-constraint}. \nonumber
\end{align}
where $\mathbf{f}_C^c$ denotes the impulses in compression phase, $\mathbf{A}$ is the inverse inertia in the contact space, and $\mathbf{v}_C^{c-}$ represents the velocity in the contact space before the compression phase. This formulation is an approximation to the Signorini condition, please see \cite{jean1999non, todorov2012mujoco} for more details. A concise derivation of \eqref{eqn:comp-cvx-prob} from the first principle is provided in Appendix~C.
\subsection{Contact model in restitution phase}
\label{sec:rest-phase}
Similarly, we can set up an optimization problem to solve for the contact impulses in the restitution phase $\mathbf{f}_C^r$. We assume the restitution follows Poisson's hypothesis, where the normal components in $\mathbf{f}_C^r$ equals those in $\mathbf{f}_C^c$ scaled by the coefficient of restitution $e_P$. We adopt Poisson's hypothesis instead of the popular Newton's hypothesis used in prior works \cite{battaglia2016interaction, de2018end, Hu2020DiffTaichi, Chen2020Symplectic}, because the latter might result in unrealistic energy increase in certain systems \cite{djerassi2009collisionA}. Please see Appendix~D for a discussion. We set up the following constraint:
\begin{align}
    \label{eqn:rest-normal-constraint}
    f_{i,n}^r \geq e_{P,i} \cdot f_{i, n}^c, \quad \forall i.
\end{align}
We have inequality instead of equality here since we would like to compensate for existing penetration in the simulation. Since we simulate the rigid body system in discrete time steps, almost every time when a collision is detected, penetration has already occurred among the objects involved in that collision. Consider the case where the collision is perfectly inelastic, i.e., COR $e_P=0$, then the true normal impulse during restitution phase would be zero, which fails to fix existing unrealistic penetration. By setting up the constraint as in Eqn. \eqref{eqn:rest-normal-constraint}, a larger normal impulse is allowed to fix existing penetration. 
% When COR is large enough such that , the equality would hold for the solution respecting Poisson's hypothesis due to the principle of maximum dissipation. 
% Thus, with Eqn. \eqref{eqn:v-jump-rest} and \eqref{eqn:rest-normal-constraint}, we take penetration compensation into account and set up the following optimization problem from the principle of maximum dissipation to solve for $\mathbf{f}_C^r$.
The optimization problem in the restitution phase is 
\begin{align}
    \label{eqn:rest-cvx-problem}
    &\underset{\mathbf{f}_C^r}{\textrm{Minimize }} \frac{1}{2} (\mathbf{f}_C^{r})^T \mathbf{A} \mathbf{f}_C^{r} + (\mathbf{f}_C^{r})^T (\mathbf{v}_C^{c+} - \mathbf{v}_C^*) \\
    &\textrm{subject to } \eqref{eqn:2nd-cone-constraint}, \eqref{eqn:non-neg-constraint}, \eqref{eqn:rest-normal-constraint}. \nonumber
\end{align}
with $\mathbf{v}_C^{c+}$ as the velocity in contact space after the compression phase and $\mathbf{v}_C^*$ as the target velocity, used for fixing penetration. If there's no penetration,  $\mathbf{v}_C^*=\mathbf{0}$. A detailed discussion of penetration compensation can be found in Appendix~G. From the principle of maximum dissipation, the equality in Eqn. \eqref{eqn:rest-normal-constraint} would hold for the solution when COR is large and penetration is small, thus respecting Poisson's hypothesis when penetration can be fixed naturally.
\subsection{Differentiability}
Solving optimization problems \eqref{eqn:comp-cvx-prob} and \eqref{eqn:rest-cvx-problem} for contact impulses allow us to calculate instantaneous velocity change to perform simulation (Algorithm \ref{alg:sim-w-c}). Moreover, we would like to back-propagate through the contact model to learn unknown properties. In fact, our proposed contact model is differentiable, thanks to recent progress on differentiable convex optimization layers. Both problems \eqref{eqn:comp-cvx-prob} and \eqref{eqn:rest-cvx-problem} are convex optimization problems with convex quadratic objectives (we show that $\mathbf{A}$ is positive semi-definite in Appendix~E) as well as linear constraints and second-order cone constraints. We can then express our problems using disciplined parametrized programming (DPP) and set up these two problems as differentiable layers using CvxpyLayers \cite{agrawal2019differentiable}. Thus, our model is differentiable and can be used in dynamics and parameter learning as well as downstream tasks.
\begin{figure*}[b]
\vspace{-1em}
    \centering
    \subfigure[]{
        \centering
        \includegraphics[width=0.13\textwidth]{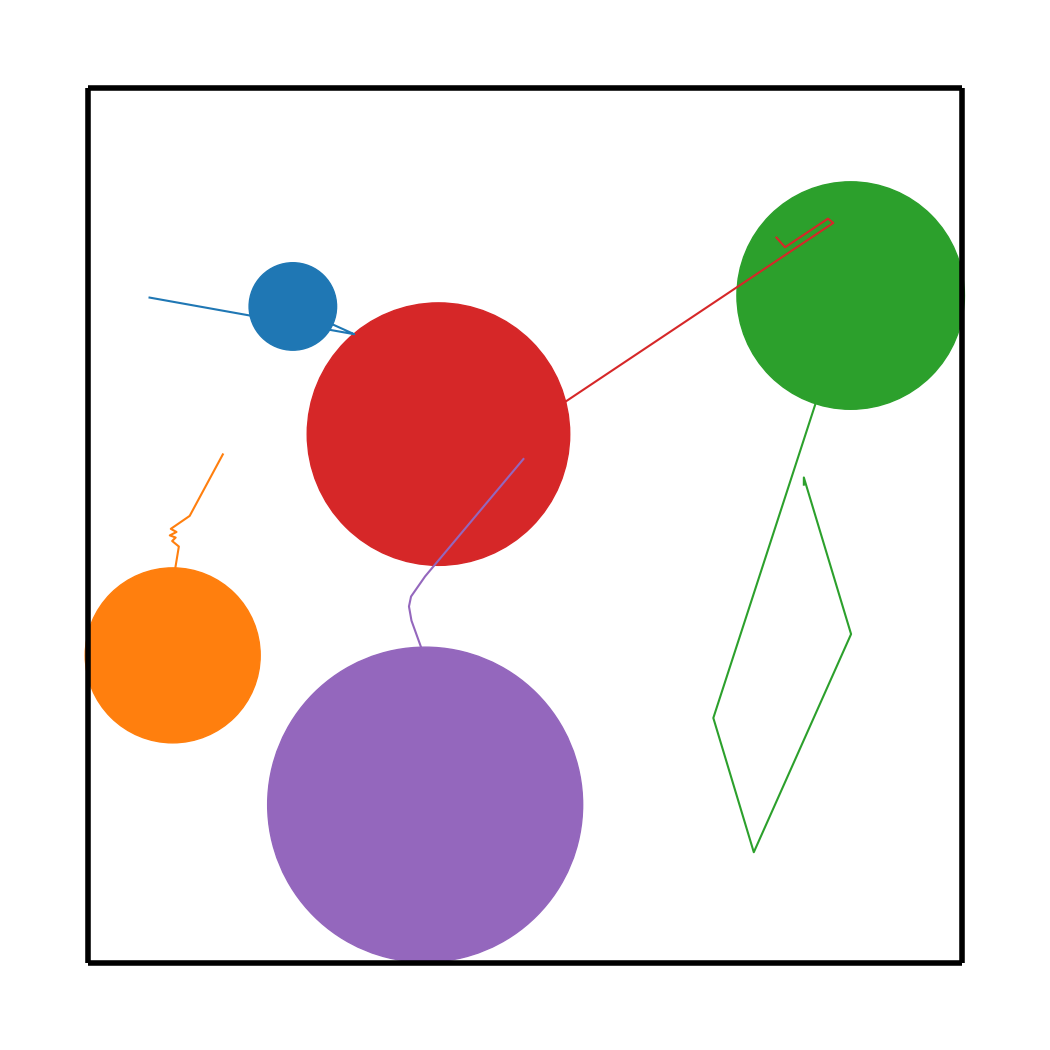}
        \label{fig:BM}
    }
    \subfigure[]{
        \centering
        \includegraphics[width=0.13\textwidth]{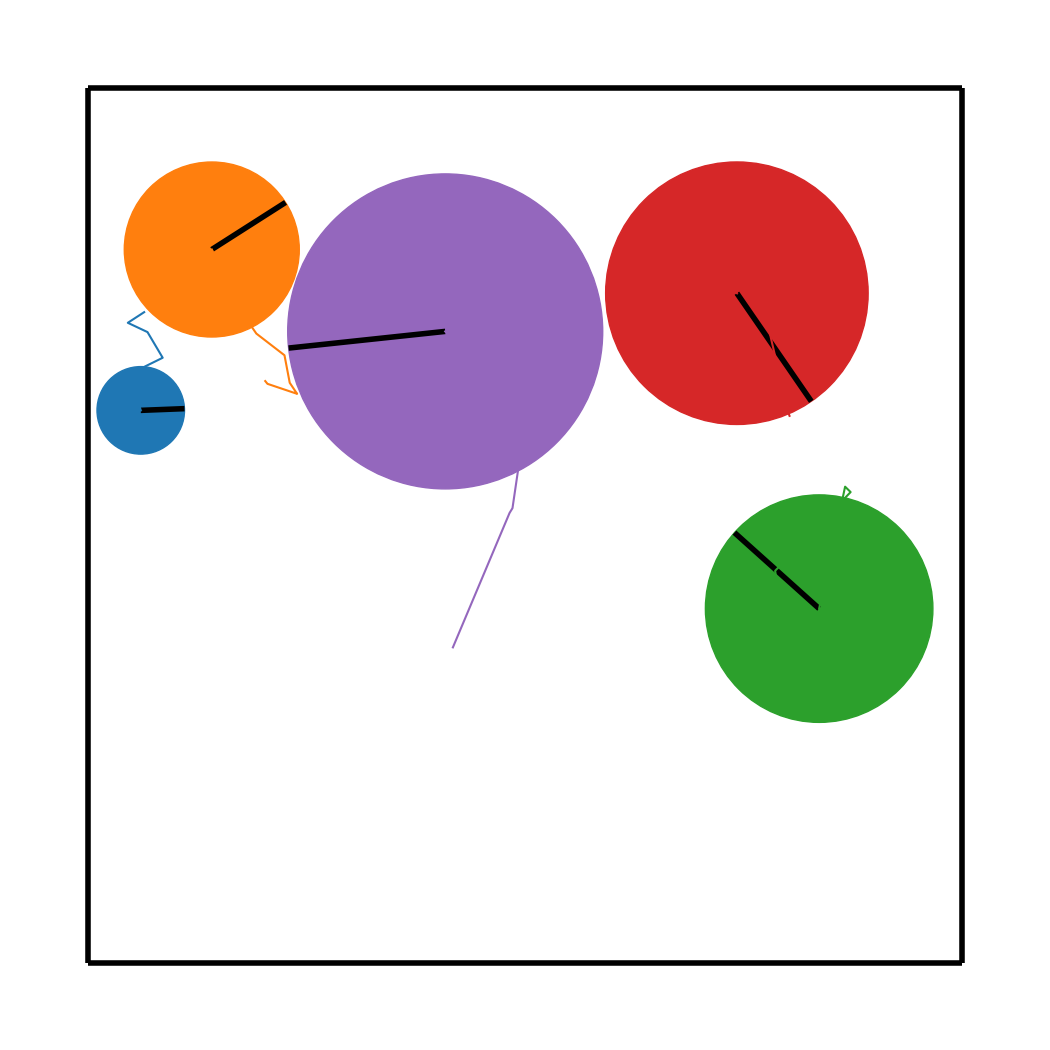}
        \label{fig:BD}
    }
    \subfigure[]{
        \centering
        \includegraphics[width=0.13\textwidth]{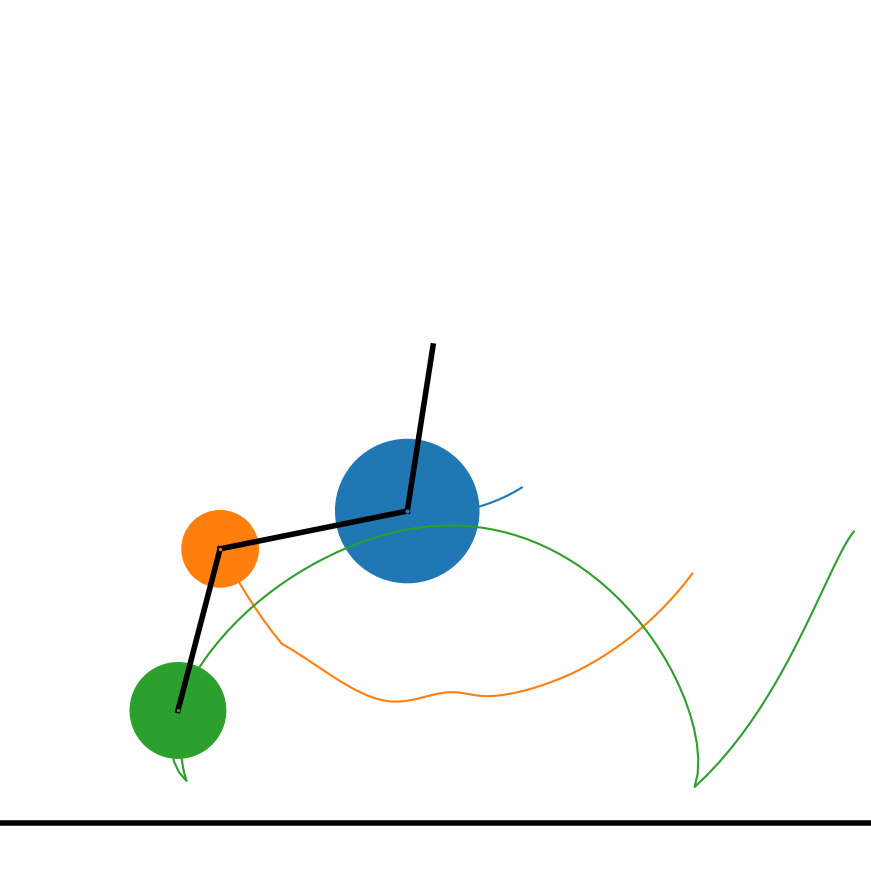}
        \label{fig:CP}
    }
    \subfigure[]{
        \centering
        \includegraphics[width=0.13\textwidth]{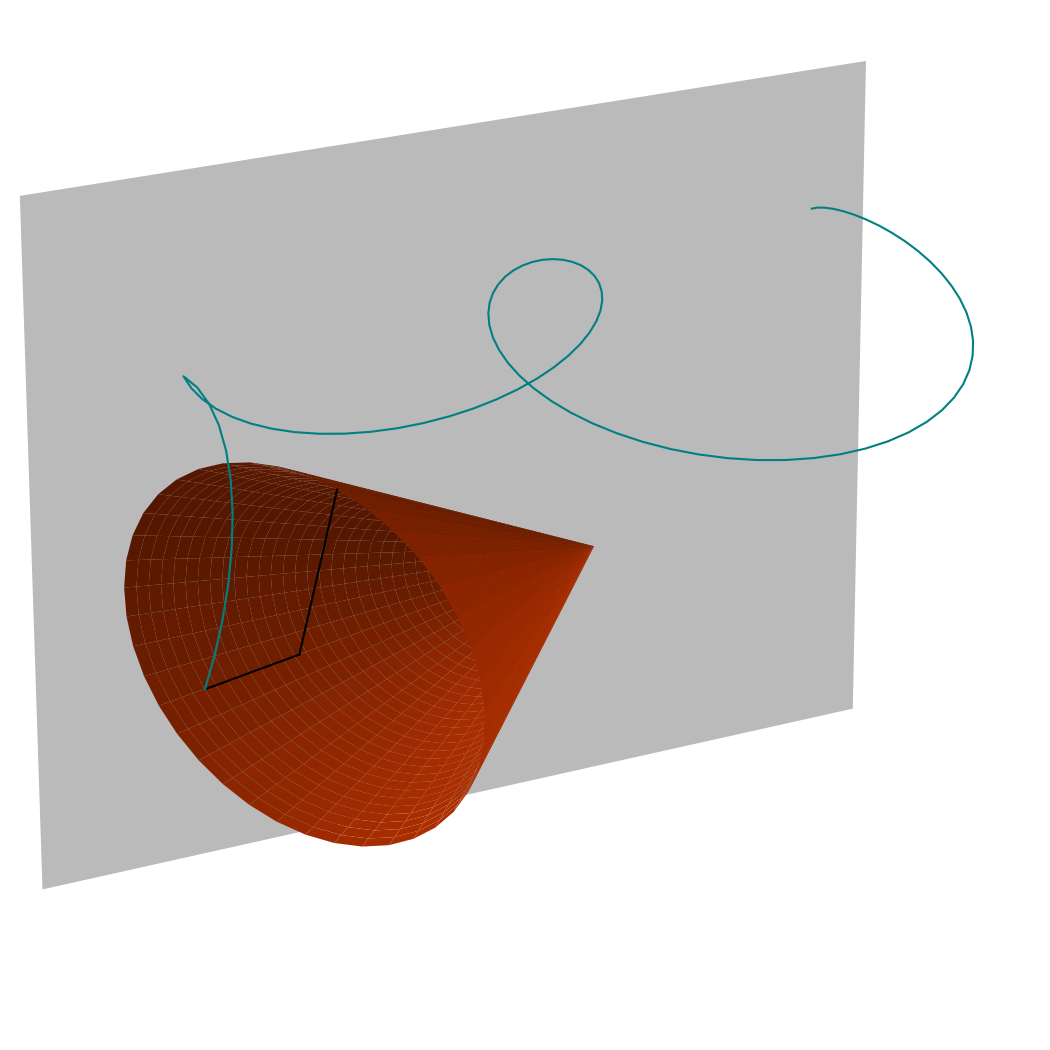}
        \label{fig:gyro}
    }
    \subfigure[]{
        \centering
        \includegraphics[width=0.13\textwidth]{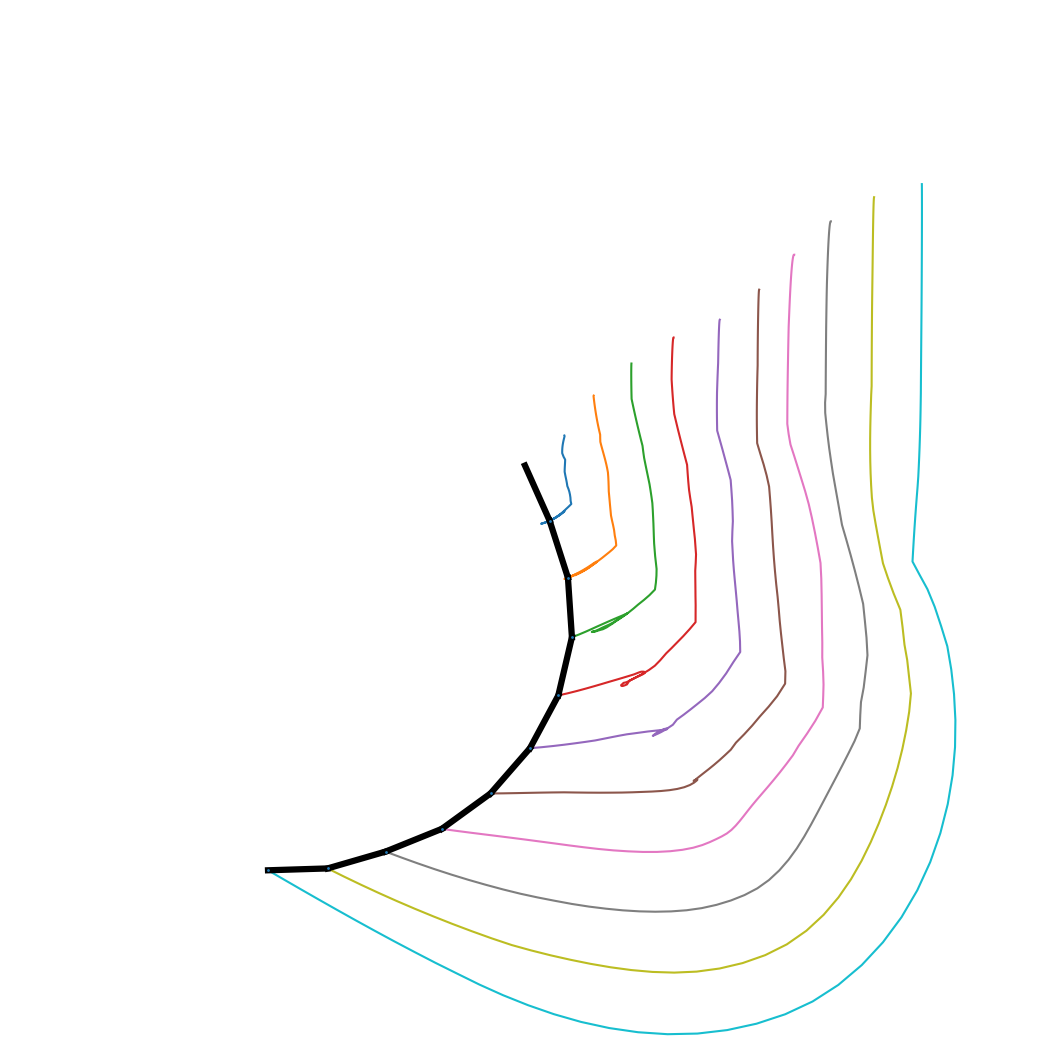}
        \label{fig:rope}
    }
    \vspace{-0.5em}
    \caption{Simulated systems with contact. From left to right: (a) bouncing point masses, (b) bouncing disks, (c) chained pendulums with ground, (d) gyroscope with a wall, and (e) rope. The black lines in bouncing disks show the orientations of disks.}
    \label{fig:systems}
\end{figure*}
\section{Experiments: dynamics and parameter learning}
\subsection{Simulated systems} 
To evaluate the proposed contact model, we simulate five different systems with contacts (Fig.~\ref{fig:systems}) and propose eight dynamics and parameter learning tasks based on these systems with different contact properties (Table~\ref{tab:system-conf}). Previous work has studied the \emph{bouncing point masses} (Fig.~\ref{fig:BM}) which is often referred to as billiards or bouncing balls \cite{Chen2020Symplectic, de2018end, Hu2020DiffTaichi}. To make this task more challenging, we let the size of each object be different. We also propose the \emph{bouncing disks} (Fig.~\ref{fig:BD}) where each disk can rotate. The 2-pendulum colliding with the ground has been used to study and analyze contact models for more than three decades \cite{kane1985dynamics}. We make this task more challenging by studying a \emph{3-Pendulum colliding with the ground} (Fig.~\ref{fig:CP}). The gyroscope is a 3D system that exhibits complex dynamics. A \emph{gyroscope colliding with a wall} (Fig.~\ref{fig:gyro}) is a system where the normal contact impulse does not point towards the center of mass (c.o.m), and Newton's hypothesis might give an unrealistic result with increased energy after collisions \cite{djerassi2009collisionA}. The \emph{rope} (Fig.~\ref{fig:rope}) has also been studied in previous works \cite{battaglia2016interaction, yang2020learning}.  Please refer to Appendix~H for further details about these systems and the tasks.
\subsection{Dynamics and parameter learning experimental setup}
\label{sec:exp-detail}
\begin{figure}[b]
    \centering
    \includegraphics[width=0.8\textwidth]{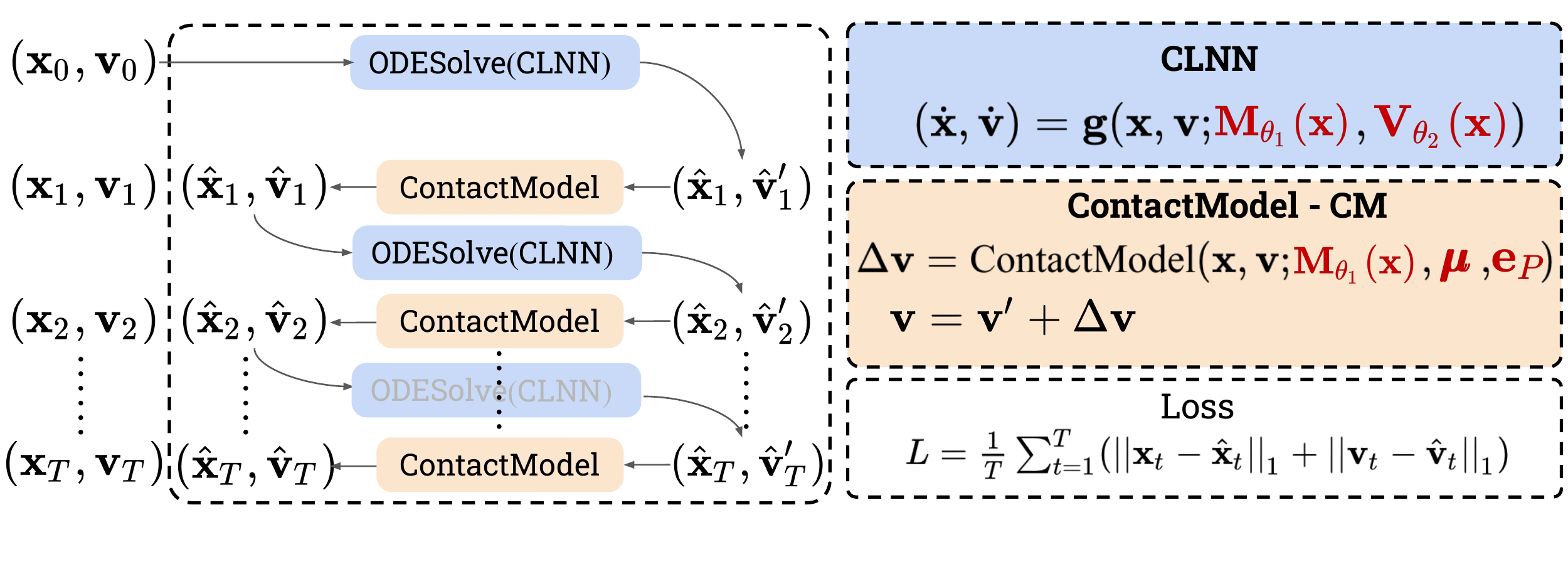}
    \caption{Dynamics and parameter learning schema of the CM-CD-CLNN model. Neural networks and learnable parameters are denoted in red. Predicted trajectories are generated using parametrized CM and CLNN. The difference between the true and predicted trajectories are minimized to learn dynamics and parameters. }
    \label{fig:schema}
\end{figure}

For each simulated system, we jointly learn system and contact properties from trajectory data by extending CLNN/CHNN with the proposed contact model. Fig.~\ref{fig:schema} shows the architecture.
% Based on these five systems, we set up 8 benchmark tasks with different contact properties to test the contact model for learning unknown contact properties. A comparison of these tasks is listed in Table \ref{tab:system-conf}. 

\textbf{Data:} For each task, the training set is generated by randomly sampling 800 collision-free initial conditions and then simulating the dynamics for 100 time steps. Since for some systems, there are very few data points in a trajectory that involves collision, we select a small chunk containing 5 consecutive time steps from each simulated trajectory such that the final training set contains 800 trajectories of length 5, where around half of the trajectories contain collisions and the other half are collision-free. We also make sure that the initial state of these selected chunks is collision-free. The evaluation and test set are generated in a similar way with 100 trajectories, respectively. 
\begin{table*}[t!]
\caption{Benchmark tasks. The columns $D$, $E$, max($C$) denote dimension of the dynamics, number of equality constraints, and the maximum number of contacts that could be simultaneously active , respectively.}
\label{tab:system-conf}
% \vskip -0.5em
\begin{small}
\begin{tabularx}{\textwidth}{c | c | c | c | c | c | c | c}
    \toprule[1pt]
    Name & System & $D$ & $E$ & max($C$) & Space & \makecell{Same $e_P$, $\mu$ for \\ all contacts} & \makecell {Conserve \\ energy} \\
    \midrule
    BP5-e & Bouncing point masses & $10$ & $0$ & $8$ & 2D & Y & Y \\
    BP5 & Bouncing point masses & $10$ & $0$ & $8$ & 2D & N & N \\
    CP3-e & Chained pendulums w/ ground & $6$ & $3$ & $1$ & 2D & Y & Y \\
    CP3 & Chained pendulums w/ ground & $6$ & $3$ & $1$ & 2D & Y & N \\
    BD5 & Bouncing disks & $30$ & $15$ & $8$ & 2D & N & N \\
    Rope & Rope & $400$ & $0$ & $\sim 399$ & 2D & Y & N \\
    Gyro-e & Gyroscope w/ a wall & $12$ & $7$ & $1$ & 3D & Y & Y \\
    Gyro & Gyroscope w/ a wall & $12$ & $7$ & $1$ & 3D & Y & N \\
    % Cloth & Cloth & 64 & 0 & 175 & 3D & Y & N \\
    \bottomrule[1pt]
\end{tabularx}
\end{small}
% \vspace{-2em}
\end{table*}

\textbf{Architecture and training details:} In the experiments, we assume the system properties, i.e., object inertia and potential energy, as well as contact properties, i.e., coefficients of friction and restitution, are unknown and need to be learned from trajectory data. The system properties are parametrized as in CLNN and CHNN \cite{finzi2020simplifying}. As for contact properties, all coefficient of friction are non-negative, so they are parametrized by scalar learnable parameters passed through \textit{ReLu} function. As each coefficient of restitution lies in the interval of $[0, 1]$, it is parametrized by a learnable parameter passed through a \textit{hard sigmoid} function. The predicted trajectories are generated by running Algorithm \ref{alg:sim-w-c} with parametrized system and contact properties. We use RK4 as the ODE solver in Neural ODE. We compute the $L_1$-norm of the difference between predicted and true trajectories, and use it as the loss function for training. The gradients are computed by differentiating through Algorithm \ref{alg:sim-w-c}, and learnable parameters are updated using the AdamW optimizer \cite{kingma2014adam, loshchilov2017decoupled} with a learning rate of 0.001.  
% Our implementation relies on publically available codebases including Pytorch \cite{NEURIPS2019_9015}, CHNN \cite{finzi2020simplifying} and Neural ODE \cite{chen2018neural}. We handle training using Pytorch Lightning \cite{falcon2019pytorch} for the purpose of reproducibility. 
%In the experiments, we assume the system properties, i.e., object inertia and potential energy, as well as contact properties, i.e., coefficients of friction and restitution, are unknown and need to be learned from trajectory data. The system properties are parametrized as in CLNN and CHNN \cite{finzi2020simplifying}. As for contact properties, each coefficient of friction is non-negative -- so it is parametrized by a learnable parameter passed through a \textit{ReLu} function. Each coefficient of restitution is in the interval of $[0, 1]$, so it is parametrized by a learnable parameter passed through a \textit{hard sigmoid} function. The predicted trajectories are generated by running Algorithm \ref{alg:sim-w-c} with parametrized system and contact properties. We use RK4 as the ODE solver in Neural ODE. We compute the $L_1$ loss of the difference between predicted and true trajectories\dz{, and use it as the loss function for training}. The gradients are computed by differentiating through Algorithm \ref{alg:sim-w-c}, and neural network parameters are updated using the AdamW optimizer \cite{kingma2014adam, loshchilov2017decoupled} with a learning rate of 0.001.

\textbf{Models:} 
We implement two slightly different versions of the contact model. The first version, referred to as CM, set up optimization problems exactly as stated in \eqref{eqn:comp-cvx-prob} and \eqref{eqn:rest-cvx-problem}. The second version, referred to as CMr, adds a diagonal positive regularization matrix $\mathbf{R}=\epsilon \mathbf{I}$ to $\mathbf{A}$ in \eqref{eqn:comp-cvx-prob} and \eqref{eqn:rest-cvx-problem}, such that $(\mathbf{A} + \mathbf{R})$ is always positive definite, which ensures a unique global minimum in each problem.\footnote{The regularization is important for obtaining the inverse dynamics, as stated in \cite{todorov2014convex}. However, the unregularized one learns more accurate dynamics and contact properties, as shown in Fig.~\ref{fig:rel_err} and Table~\ref{tab:leaned-cp}.} These two versions are combined with CLNN and CHNN to set up the following four neural network models: (\textit{i}) CM-CD-CLNN, (\textit{ii}) CM-CD-CHNN, (\textit{iii}) CMr-CD-CLNN, and (\textit{iv}) CMr-CD-CHNN. The ``CD" in model names emphasizes that we assume that a collision detection module is given.
\begin{figure}[b]
    \centering
    \includegraphics[width=1.0\textwidth]{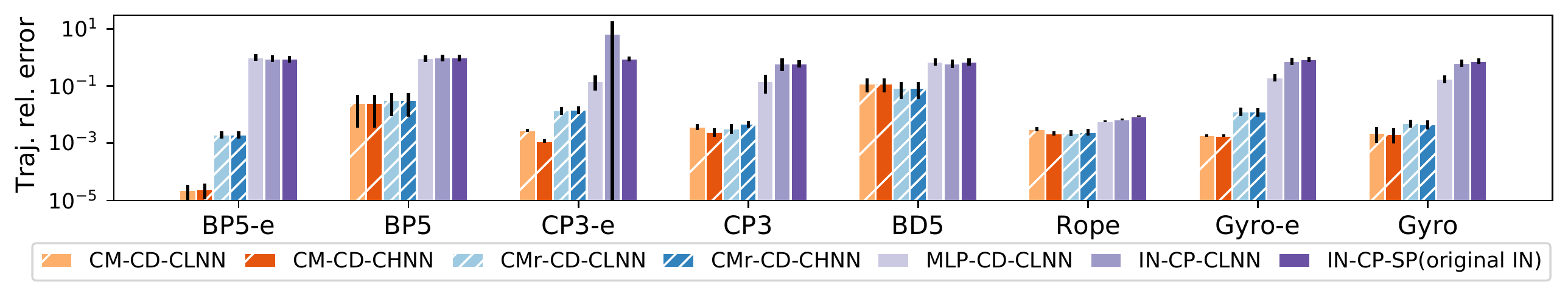}
    \caption{Trajectory relative error (log scale) with 95\% confidence interval error bars. Each error is averaged over 100 test trajectories of length 5. }
    \label{fig:traj_rel_err}
\end{figure}
\begin{figure}[t]
    \centering
    \includegraphics[width=0.99\textwidth]{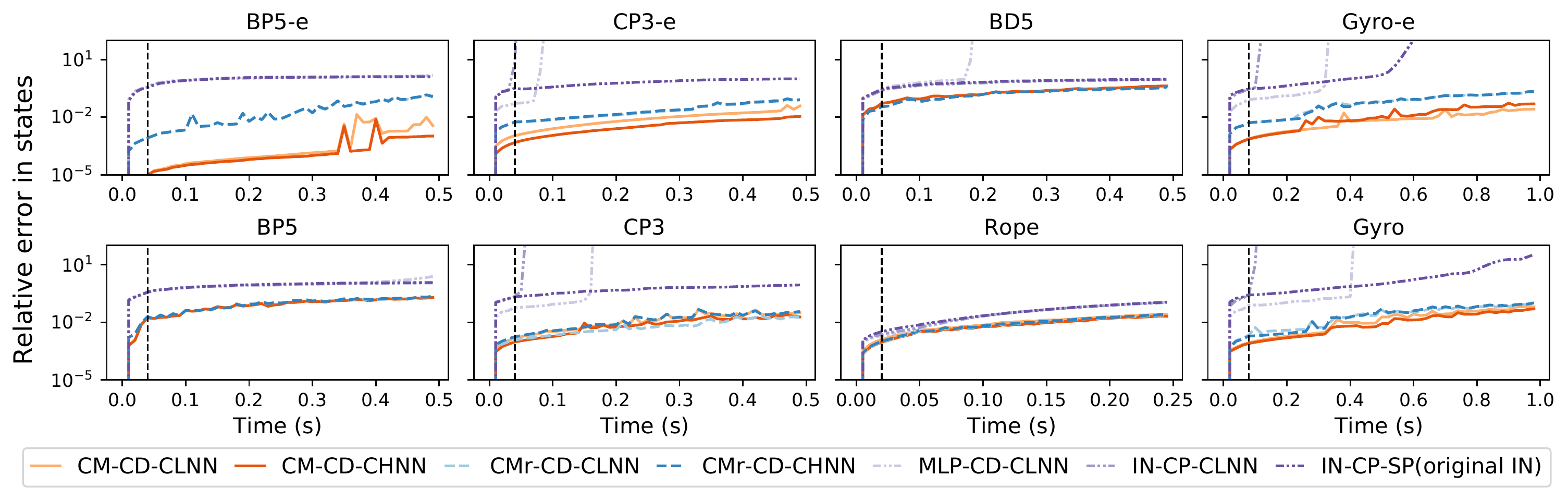}
    \caption{Relative error (log scale) along long test trajectories (50 times steps). Each curve is averaged over 100 test trajectories. Vertical dashed lines show the trajectory length during training.}
    \label{fig:rel_err}
\end{figure}

\textbf{Baselines:}
We also set up three baselines. In the first baseline, MLP-CD-CLNN, we calculate the instantaneous velocity change from a multi-layer perceptron (MLP) instead of the proposed contact model. Our second baseline, IN-CP-CLNN, calculates velocity change from an interaction network (IN) \cite{battaglia2016interaction} without requiring a collision detection module, since IN has the ability to learn collisions and contact. IN requires system and contact properties as input. Here the ``CP" in the model name emphasizes true contact properties are given and the system properties learned by CLNN are fed into IN. Our last baseline, IN-CP-SP, is the original interaction network which has shown strong ability in predicting 2D rigid body trajectories without equality constraints, but haven't been tested on systems with equality constraints or 3D systems. The name emphasizes that true system and contact properties are known and are fed into IN. Also, the name indicates no collision detection module is needed in this baseline. To train these baseline models, we transform each trajectory into multiple one-step pairs, as has been done in IN \citep{battaglia2016interaction}. We also attempted to use the LCP formulation of contact model \cite{de2018end} as a baseline. However, the implementation of gradient computation of the LCP function in \cite{de2018end} results in NaN in our examples. Please refer to Appendix~J for additional details. As the forward computation of LCP works as expected, we use LCP-generated training data to test the robustness of our model.
%For baseline models, we transform each trajectory into multiple one-step pairs, as done in IN.
%
%
%
%
\subsection{Dynamics and parameter learning results}
\label{sec:result}
Our implementation relies on publicly available codebases including Pytorch \cite{NEURIPS2019_9015}, CHNN \cite{finzi2020simplifying}, Symplectic ODE-Net \cite{Zhong2020Symplectic} and Neural ODE \cite{chen2018neural}. We handle training using Pytorch Lightning \cite{falcon2019pytorch} for the purpose of reproducibility. 
% All data generation and training are done using a Intel Xeon CPU. 

\textbf{Prediction:} We report the average relative $L_1$ error over the test trajectories of 7 models on 8 tasks in Fig. \ref{fig:traj_rel_err}. In all tasks, our models beat baseline models. The performance difference between CLNN and CHNN is minor since their architectures are similar. In most tasks, CM outperforms CMr. In the BP5-e task, CM beats CMr by 2 orders of magnitude. 
% \bd{In two 3D gyroscope tasks, CM performs badly, while CMr performs the best.} The reason is discussed below. 
%
%
% mainly because CM fails to learn the true contact properties, as shown in Table \ref{tab:leaned-cp}. On the contrary, CMr learns contact properties that is close to the ground truth, explaining its good performance in predicting gyroscope tasks.
IN does not perform well even in BP5 tasks since our training set (3.2k one-step pairs) is much smaller than the dataset (1M one-step pairs) used in the IN paper. We also report average relative $L_1$ errors along test trajectories of 50 time steps in Fig.~\ref{fig:rel_err}, in order to show each model's ability in long term prediction. We observe that our contact models CM and CMr outperform baselines in all tasks.
% 2D tasks and CMr outperforms baselines in 3D tasks. 

%
%
\textbf{Interpretable mass ratio:} Without direct supervision on mass, deep learning algorithms are unlikely to recover the true mass, as pointed out in \citep{Zhong2020Symplectic}. However, one can still inspect the ratio of learned mass values to see how well this physical property is learned. The mass ratio plays an important role in determining the motion of objects when they interact with each other, e.g., during collisions. In our BP5 task, CM-CD-CLNN learns the mass ratio $[m_2/m_1, m_3/m_1, m_4/m_1, m_5/m_1] = [2.0001, 6.0036, 8.0014, 10.0024]$, which is very close to the true ratio $[2, 6, 8, 10]$. In fact, our framework is able to accurately learn mass ratios across tasks (please see Appendix~I for details).
\begin{table}[b]
\vspace{-5pt}
\caption{Learned contact properties from our 4 models on 6 tasks that has unique contact properties for all contacts. Bold numbers are the best learned contact properties in each task across 4 models.}
\label{tab:leaned-cp}
\vskip 0.1in
\begin{small}
\begin{tabularx}{\textwidth}{c | Y | Y | Y | Y | Y | Y | Y| Y | Y | Y | Y | Y}
    \toprule[1pt]
     & \multicolumn{2}{c|}{BP5-e} & \multicolumn{2}{c|}{CP3-e} & \multicolumn{2}{c|}{CP3} & \multicolumn{2}{c|}{Rope} &
     \multicolumn{2}{c|}{Gyro-e} &
     \multicolumn{2}{c}{Gyro} \\
     & $\mu$ & $e_P$ & $\mu$ & $e_P$ & $\mu$ & $e_P$ & $\mu$ & $e_P$ & $\mu$ & $e_P$ & $\mu$ & $e_P$ \\
     \midrule[0.5pt]
     Ground Truth &  \makebox[10pt]{0.000} & \makebox[10pt]{1.000} & \makebox[10pt]{0.000} & \makebox[10pt]{1.000} & \makebox[10pt]{0.500} & \makebox[10pt]{0.000} & \makebox[10pt]{0.000} & \makebox[10pt]{0.000} & \makebox[10pt]{0.000} & \makebox[10pt]{1.000} & \makebox[10pt]{0.100} & \makebox[10pt]{0.800}\\
     \midrule[0.5pt]
     CM-CD-CLNN & \makebox[10pt]{\textbf{0.000}} & \makebox[10pt]{\textbf{1.000}} & \makebox[10pt]{\textbf{0.000}} & \makebox[10pt]{\textbf{1.000}} & \makebox[10pt]{0.500} & \makebox[10pt]{0.005} & \makebox[10pt]{0.026} & \makebox[10pt]{0.000} & \makebox[10pt]{\textbf{0.000}} & \makebox[10pt]{\textbf{1.000}} & \makebox[10pt]{\textbf{0.100}} & \makebox[10pt]{\textbf{0.800}}\\
     CM-CD-CHNN & \makebox[10pt]{\textbf{0.000}} & \makebox[10pt]{\textbf{1.000}} & \makebox[10pt]{\textbf{0.000}} & \makebox[10pt]{\textbf{1.000}} & \makebox[10pt]{\textbf{0.500}} & \makebox[10pt]{\textbf{0.004}} & \makebox[10pt]{\textbf{0.017}} & \makebox[10pt]{\textbf{0.000}} & \makebox[10pt]{\textbf{0.000}} & \makebox[10pt]{\textbf{1.000}} & \makebox[10pt]{\textbf{0.100}} & \makebox[10pt]{\textbf{0.800}}\\
     CMr-CD-CLNN & \makebox[10pt]{\textbf{0.000}} & \makebox[10pt]{\textbf{1.000}} & \makebox[10pt]{0.002} & \makebox[10pt]{1.000} & \makebox[10pt]{0.500} & \makebox[10pt]{0.023} & \makebox[10pt]{0.037} & \makebox[10pt]{0.011} & \makebox[10pt]{0.002} & \makebox[10pt]{1.000} & \makebox[10pt]{0.099} & \makebox[10pt]{0.892}\\
     CMr-CD-CHNN & \makebox[10pt]{\textbf{0.000}} & \makebox[10pt]{\textbf{1.000}} & \makebox[10pt]{0.002} & \makebox[10pt]{1.000} & \makebox[10pt]{0.500} & \makebox[10pt]{0.023} & \makebox[10pt]{0.046} & \makebox[10pt]{0.019} & \makebox[10pt]{0.002} & \makebox[10pt]{1.000} & \makebox[10pt]{0.099} & \makebox[10pt]{0.893}\\
    \bottomrule[1pt]
\end{tabularx}
\end{small}
\end{table}

\textbf{Interpretable contact properties:} Table \ref{tab:leaned-cp} shows the learned contact properties by our 4 models in 6 tasks where the contact properties are the same for all contacts. For all tasks, CM can learn contact properties accurately which explains its good performance in prediction. CMr is an approximate model and does not infer contact properties as accurately as CM. The interpretability of the learned contact properties along with the mass ratio explains the performance of our framework and shows that the proposed contact model indeed extends Lagrangian and Hamiltonian neural networks.
% For 3D tasks, CM fails to infer contact properties, which explains the large prediction error in Figure \ref{fig:traj_rel_err}. This can probably be attributed to the fact that the positive semi-definiteness of matrix $\mathbf{A}$ can introduce some errors in gradient computation. By using a positive semi-definite matrix $(\mathbf{A} + \mathbf{R})$, CMr avoids this problem and infers contact properties that are close to the ground truth. 
%\dz{The interpretability of the learned contact properties along with mass ratio explains the performance of our framework and shows that the proposed contact models indeed extend Lagrangian and Hamiltonian neural networks.}
\begin{figure}[t!]
    \centering
    \includegraphics[width=0.99\textwidth]{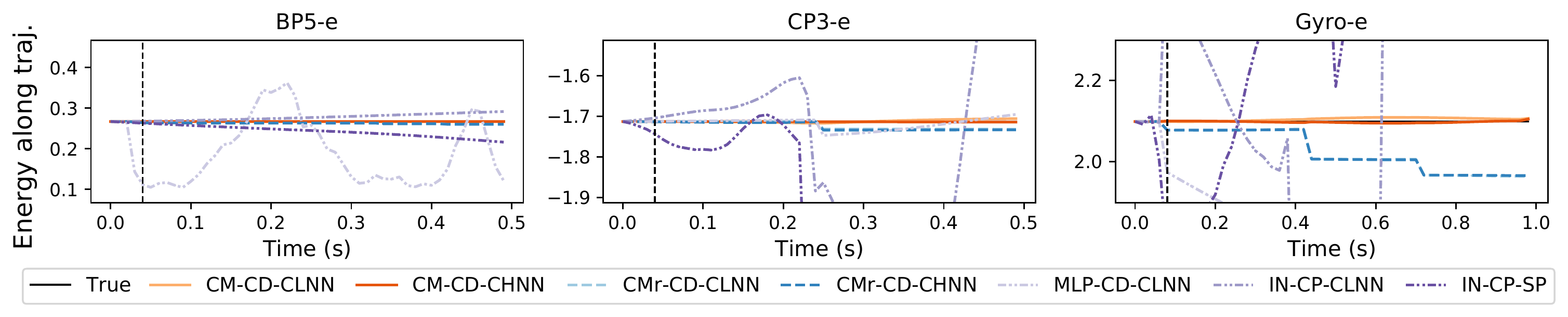}
    \vspace{-5pt}
    \caption{Energy of the predicted trajectories of all 7 models on a sampled test initial condition from BP5-e, CP3-e and Gyro-e tasks. The true energy in each task is represented by the horizontal black line in the middle, which is constant along the trajecotry.}
    \label{fig:traj_energy}
\end{figure}

\textbf{Energy:} The prior of Lagrangian/Hamiltonian dynamics conserve energy along each collision-free trajectory, which is one of the reason that Lagrangian/Hamiltonian-based neural network models perform better in prediction and generalization \cite{ lutter2018deep, greydanus2019hamiltonian, Zhong2020Symplectic,cranmer2020lagrangian}. Fig.~\ref{fig:traj_energy} illustrates how the total energy changes over time for the predicted trajectories of 7 models on 3 tasks that conserve energy since the contacts are elastic and frictionless (i.e., $e_P=1, \mu=0$). Models using CM perform the best in conserving energy in all three tasks since CM learns contact properties perfectly (Table~\ref{tab:leaned-cp}). This demonstrates that the proposed contact model can uncover the energy conserving aspect even though energy conservation has not been enforced explicitly. For CP3-e and Gyro-e systems, models using CMr lose energy each time collision happens  since they learn positive coefficients of friction in these tasks (Table \ref{tab:leaned-cp}). The baseline models perform the worst in terms of energy conservation.
% The prior of Lagrangian/Hamiltonian dynamics conserve energy along each trajectory, which is one of the reason that Lagrangian/Hamiltonian-based neural network models perform better in prediction and generalization. An elastic frictionless contact ($e_P=1, \mu=0$) also conserves energy. Fig.~\ref{fig:traj_energy} illustrates how the total energy changes over time for the predicted trajectories of 7 models on three tasks that conserves energy. Models using CM conserves energy in all three tasks since CM learns contact properties perfectly (Table \ref{tab:leaned-cp}). This shows that the proposed contact model can uncover the energy conserving aspect even though energy conservation has not been not enforced explicitly. Models using CMr lose energy each time collision happens in CP3-e and Gyro-e, since they learn positive coefficients of friction (Table \ref{tab:leaned-cp}). The baseline models perform the worst in terms of energy conservation.
%

\begin{wrapfigure}[17]{r}{0.35\linewidth}
\begin{minipage}[t]{1\linewidth}
% \vspace{-4pt}
% \begin{figure}[H]% {r}{0.35\textwidth}
    % \vspace{-15pt}
    % \centering
    \includegraphics[width=1.0\linewidth]{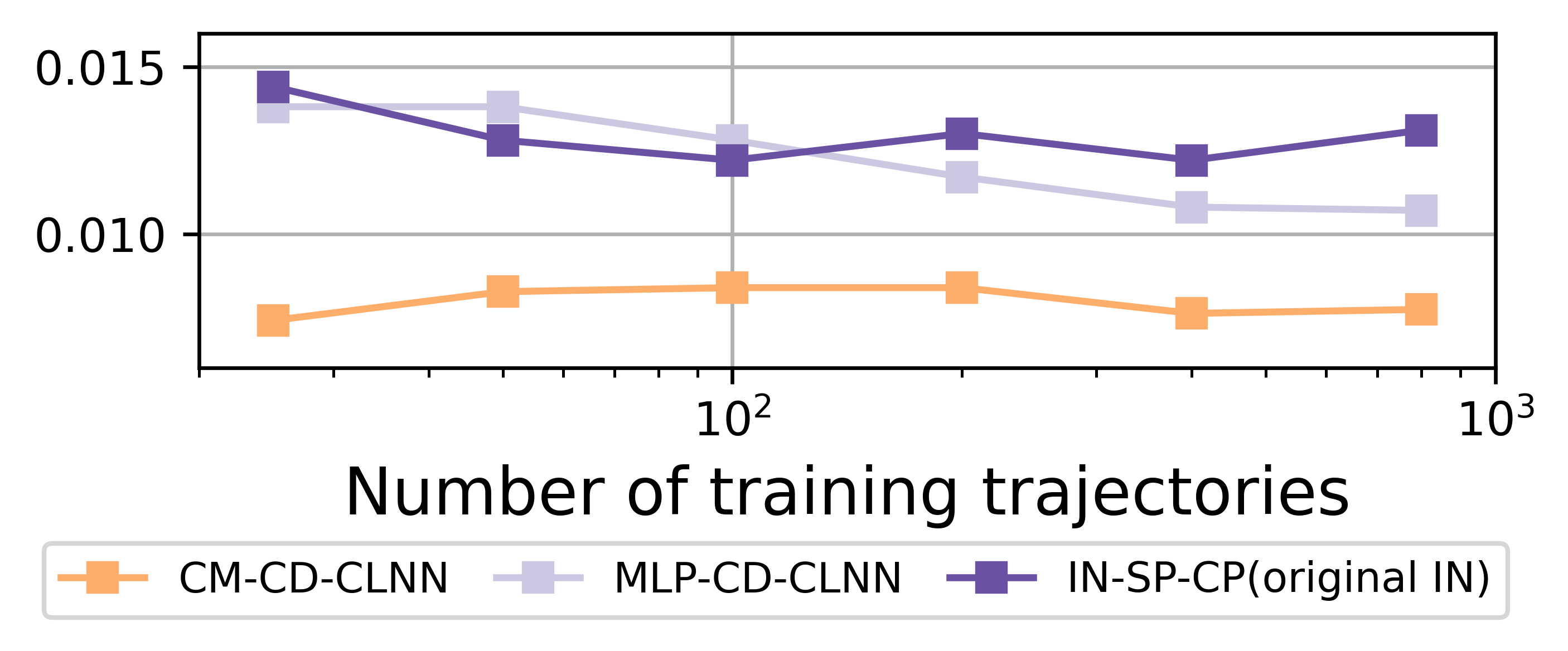}
    \captionof{figure}{Validation losses for the BP5 task.}
    \label{fig:sample_efficiency}
% \end{figure}

% \begin{table}%{r}{0.33\textwidth}
\captionof{table}{Average wall clock time in each iteration of ropes. The last column shows increases in time}
\label{tab:scalability}
% \vskip 0.1in
\begin{small}
\begin{tabular}{c|c|c|c}
    \toprule[1pt]
    $D$ & max($C$) & time (s)&  \\
    \midrule
    100 & $\sim 99$ & 0.869 & 1x \\
    200 & $\sim 199$ & 1.563 & 1.7x \\
    400 & $\sim 399$ & 3.225 & 3.7x \\
    \bottomrule[1pt]
\end{tabular}
\end{small}
% \end{table}
\end{minipage}
\end{wrapfigure}

%
%
%
\textbf{Sample efficiency:} We use the BP5 task to demonstrate the sample efficiency of this proposed framework. We vary the training sample size from 25 to 800 trajectories and report the validation loss ($L_1$-norm) of CM-CD-CLNN, MLP-CD-CLNN, and Interaction Network. Figure~\ref{fig:sample_efficiency} shows that our framework works well even with limited training data. 

\textbf{Scalability:} In Table~\ref{tab:scalability}, we report the average wall clock time in each iteration (forward pass and backward pass) during training of three sizes of ropes. The time scales approximately linearly with the numbers of coordinates ($D$) and contacts ($C$). See Appendix L for additional results for scalability.

\textbf{Robustness:} We evaluate the robustness of our framework by training our model using data generated by LCP formulation and noisy data. (See Appendix~K for details.) When trained on LCP data, our framework can learn accurate contact properties in 2D tasks. For the 3D Gyro tasks, the learned contact properties are not as accurate (e.g., learned COR of $0.822$ instead of $0.800$). This is expected since the LCP formulation relaxes the 3D friction cone into a (linear) polyhedral cone and the direction of friction impulses would deviate from those given by our contact model, which is based on the second-order friction cone. In addition, we observe that the performance of our model does not suffer from noisy data since we incorporate strong physics prior into deep learning. For CMr, we also perform an ablation test to investigate the influence of the amount of regularization. By setting the regularizer as $\mathbf{R}=\epsilon \mathbf{I}$, we observe that a smaller $\epsilon$ (e.g. 0.001) result in more accurate learned contact properties. This is expected since a smaller $\epsilon$ approximate \eqref{eqn:comp-cvx-prob} and \eqref{eqn:rest-cvx-problem} better. However, making $\epsilon$ a learnable parameter does not improve accuracy. Please see Appendix~K for more details.
% Our framework also perform well when trained on noisy data. 
%For CMr, we also perform an ablation test on the regularizer $\mathbf{R}=\epsilon \mathbf{I}$.
%We also make $\epsilon$ learnable but find it does not lead to better performance.

%
%
\section{Experiments: downstream tasks}
Since our framework is differentiable, we can use it as a differentiable physics simulator to solve downstream tasks after we have learned the system and contact properties. Here we demonstrate this capability by considering three gradient-based trajectory planning tasks and using CM-CD-CLNN.
%Since our framework is differentiable, after learning system and contact properties, we can use the learned framework as a differentiable physics simulator to solve downstream tasks. Here we demonstrate three gradient-based trajectory planning tasks using CM-CD-CLNN.

\begin{figure}[b]
    \centering
    \subfigure[]{
        \centering
        \includegraphics[width=0.33\textwidth]{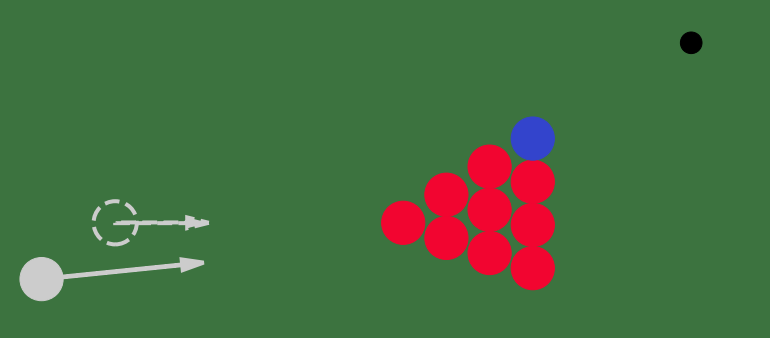}
        \label{fig:billiards_our_model}
    }
    % \hspace{-20pt}
    \subfigure[]{
        \centering
        \includegraphics[width=0.33\textwidth]{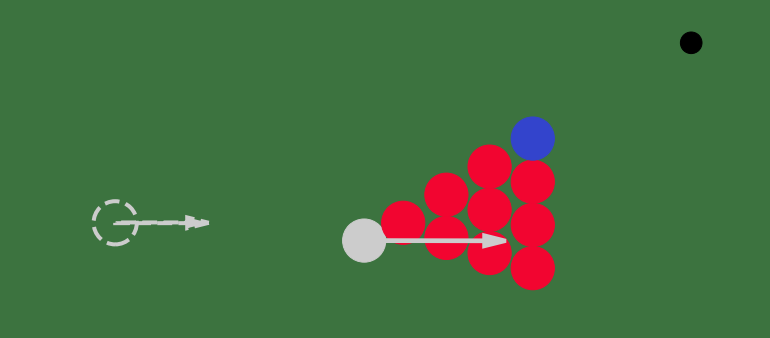}
        \label{fig:billiards_difftaichi}
    }
    % \hspace{-20pt}
    \subfigure[]{
        \centering
        \includegraphics[width=0.27\textwidth]{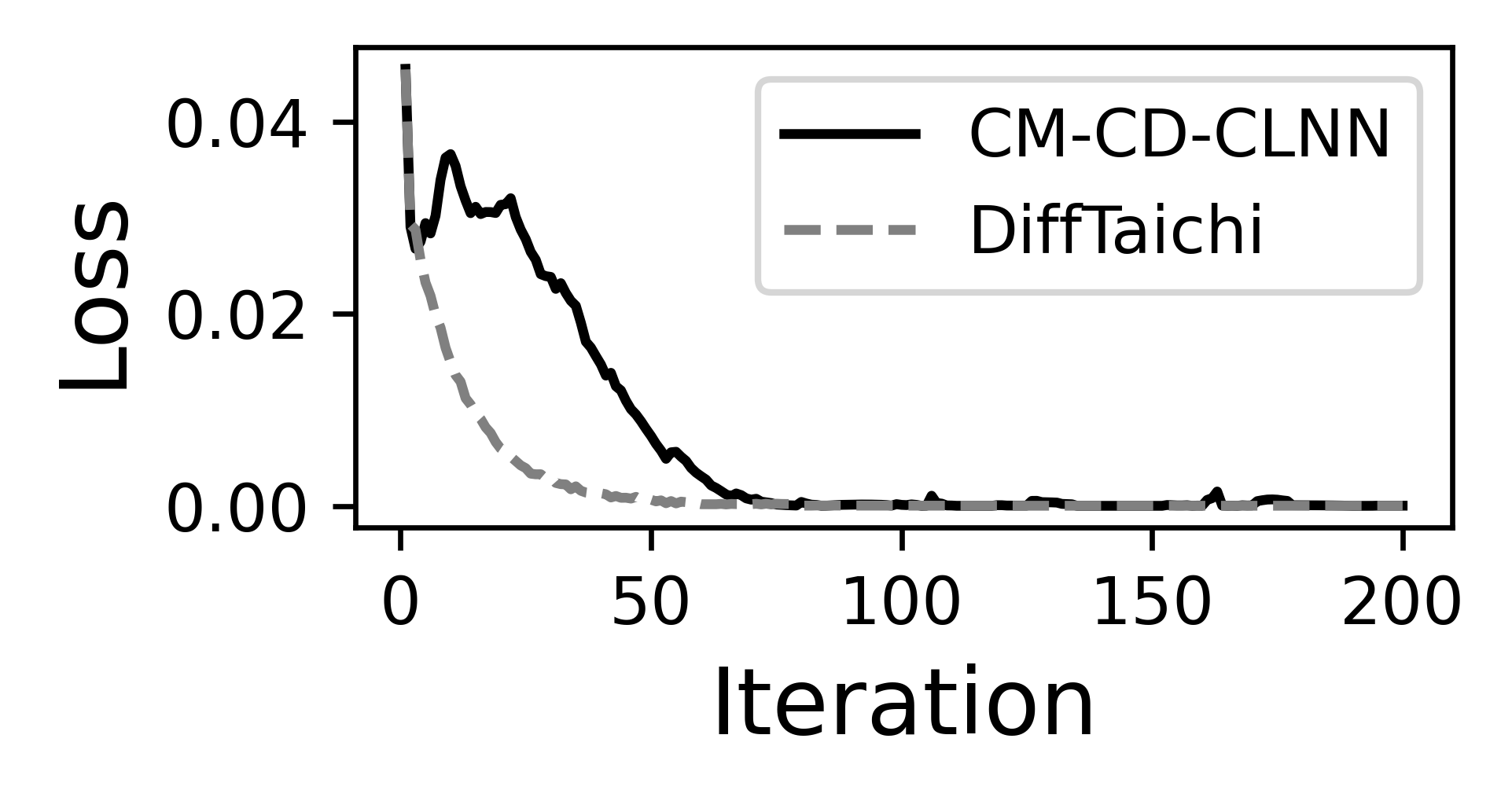}
        \label{fig:billiards_loss}
    }
    \vspace{-8pt}
    \caption{Billiards. (a), (b): The solid white ball and arrow shows the initial position and velocity optimized by CM-CD-CLNN and DiffTaichi, respectively, while dashed white ball and arrow shows those of the white ball before optimization. (c): Loss as a function of training iterations.}
    \label{fig:billiards}
\end{figure}

\textbf{Billiards:} We study the same billiard task as in DiffTaichi \cite{Hu2020DiffTaichi}. The goal is to find the initial position and velocity of the white ball such that blue ball hit the black target at the 1024th time step. In order to test our framework's ability to solve downstream task and make comparison with DiffTaichi, we assume the parameters such as mass and contact properties are known, the same assumption in DiffTaichi. Fig~\ref{fig:billiards_our_model} and \ref{fig:billiards_difftaichi} shows the solution found by our proposed model and DiffTaichi, respectively. This task does not have a unique solution since one can place the white ball closer to the billiards with a relatively small initial velocity (e.g. DiffTaichi solution) or place the ball farther away from the billiards with a relatively large initial velocity (e.g. our solution). Fig~\ref{fig:billiards_loss} compares the convergence, where the loss is the distance between the black target and the blue ball at the 1024th time step. DiffTaichi has better convergence probably because it implements a simpler contact and dynamics model and it takes time of impact (TOI) into account. The TOI might be able to explain why the optimized positions of the white balls in DiffTaichi and our method are on the right and left of the initial guess, respectively - the gradient w.r.t. the initial position using naive integrator and TOI have different signs (Figure 4 in \cite{Hu2020DiffTaichi}).
% Since in this example DiffTaichi implements a simpler contact and dynamics model, it has better convergence in learning, as shown in Fig~\ref{fig:billiards_loss}, where loss is the distance between the black target and the blue ball after 1024 time steps. 

\begin{wrapfigure}[14]{r}{0.35\textwidth}
    \vspace{-6pt}
    \centering
    \subfigure[]{
        \centering
        \includegraphics[width=0.18\textwidth]{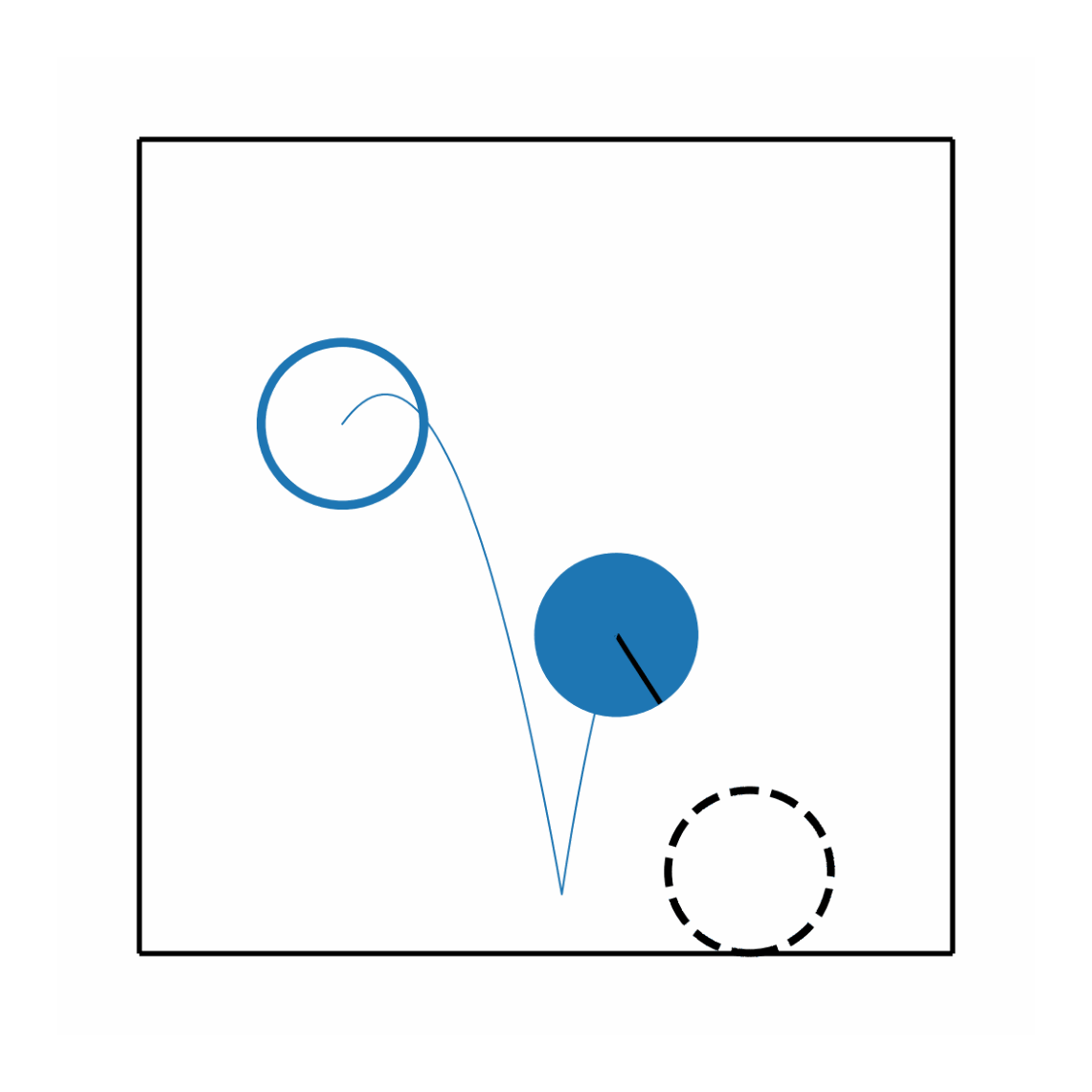}
        \label{fig:hit}
    }
    \hspace{-20pt}
    \subfigure[]{
        \centering
        \includegraphics[width=0.18\textwidth]{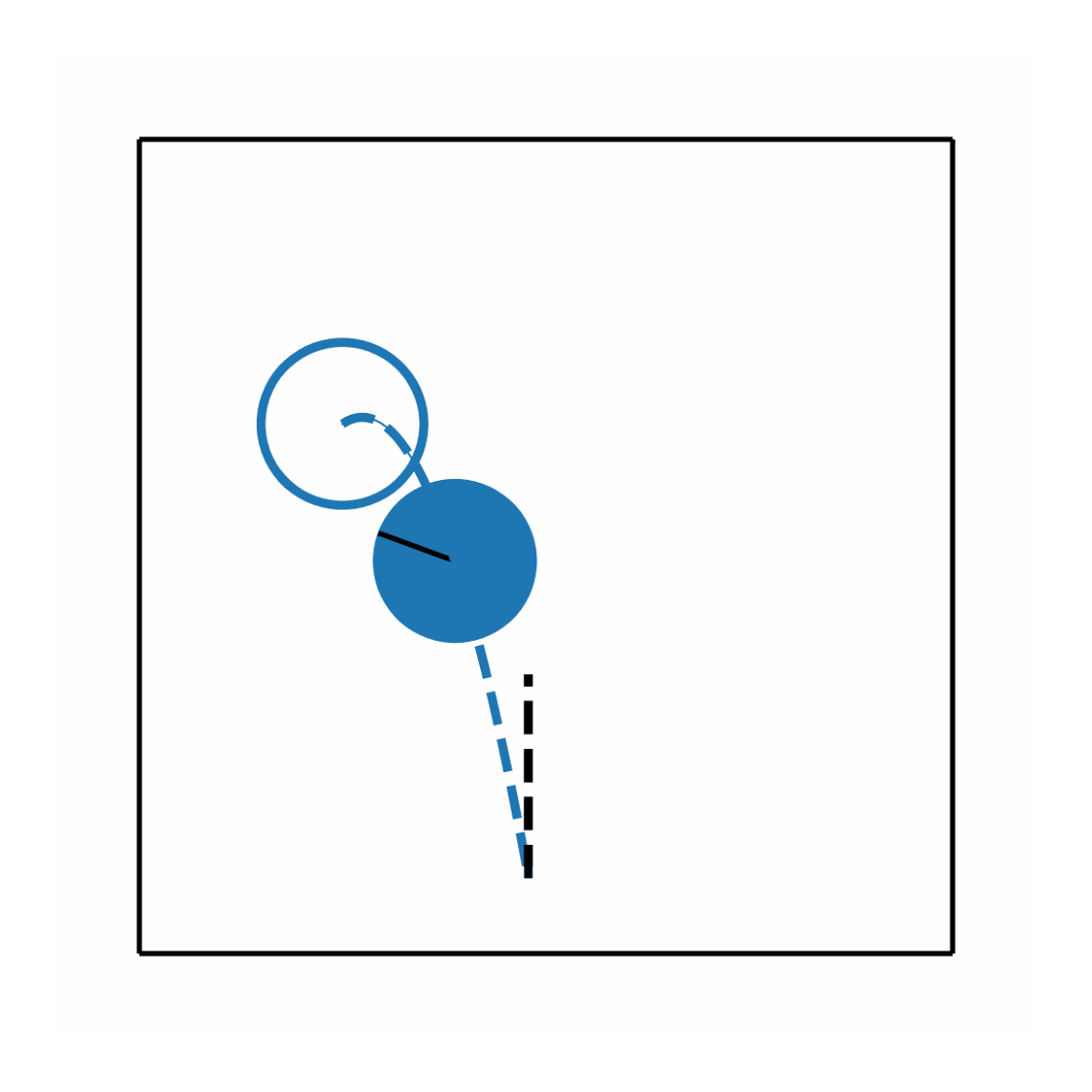}
        \label{fig:vertical}
    }
    % \vspace{-10pt}
    \caption{Blue hollow circles indicate the initial position of the disk. (a) the ``hit'' task. The black hollow circle indicate the target position. (b) the ``vertical'' task. }
    \label{fig:throw}
    \vspace{-5pt}
\end{wrapfigure}

\textbf{Throwing:} We present two throwing tasks as shown in Fig.~\ref{fig:throw}. These throwing tasks are simplified versions of similar tasks studied in  \cite{10.1145/3414685.3417766, 10.1111/cgf.14104}, but we solve the tasks based on learned dynamics while previous works \cite{10.1145/3414685.3417766, 10.1111/cgf.14104} solve them with true dynamics.  In the ``hit'' task (Fig. \ref{fig:hit}), 
the initial position of the disk is fixed, the goal is to find a desired initial velocity so that the disk reach the target (black circle) after exactly one bounce off the ground. 
In the ``vertical'' task (Fig. \ref{fig:vertical}), the initial position and the translational velocity are fixed, so that first half of the center of mass (c.o.m) trajectory (dashed blue) is fixed. The goal is to find a desired initial angular velocity such that the second half of the c.o.m trajectory is as close to a vertical line (dashed black) as possible. In this task we need to learn a counter-clockwise spin such that when the disk bounces off the ground, there are enough friction to stop the horizontal motion.

For these two tasks, we parametrize the initial condition to be learned, simulate the trajectory based on the learned system and contact properties, and minimize the difference between simulated outcome and the goal by gradient descent. We can successfully find the initial conditions to achieve the tasks, evaluated using the true system and contact properties. (Please see the video for additional details).

\section{Conclusion}
\label{sec:conclusion}
In this work, we have introduced a differentiable contact model, which can capture contact dynamics with different properties. Our contact model extends the applicability of Lagrangian/Hamiltonian-inspired neural networks to enable the learning of hybrid dynamics in rigid body systems and offer interpretability about system and contact properties. We show that the proposed framework achieves better prediction with fewer samples and is robust against noisy data or LCP-generated data. Future works will incorporate model-based control and explore interpretable safe control policies for robotics applications. A particular direction could be to develop appropriate energy shaping control policies~\citep{9029624} and integrate them with this proposed learning framework.

\textbf{Limitations:} 
Our framework assumes a known collision detection module. Although it can be obtained from an idealized touch feedback sensor \cite{pmlr-v130-hochlehnert21a}, this information might be unavailable in other scenarios. Future work would explore how to relax this assumption.
Our framework might fail to correctly simulate systems which have extremely high mass ratios or stiffness ratios as compared to \citet{10.1111/cgf.14104}, where they show their primal method and dual method perform well in high mass ratios and high stiffness ratios scenarios, respectively. Our model might also have challenges in contact-rich systems and might not be as scalable as IPC \cite{li2020incremental}. Please see Appendix L for additional scalability results. Our framework also uses a mix of acceleration-based simulation (integrating continuous dynamics) and time-stepping methods (calculating instantaneous velocity change) while other simulation methods typically use only one of them. This is because we'd like to use RK4 to better enforce the conservation of energy as done in \cite{Zhong2020Symplectic, finzi2020simplifying}. However, this choice also makes our simulator not as efficient as other simulators. We would like to compare our method to other differentiable physics model such as NeuralSim~\cite{NeuralSim_icra21} and gradSim~\cite{murthy2021gradsim}. However, gradSim has not been open sourced when this work is conducted and it is hard to reproduce model.  NeuralSim has its own automatic differentiation engine where gradient are computed one at the time, which is suitable for downstream tasks as demonstrated in \cite{NeuralSim_icra21}; however, it is not suitable for dynamics and parameter learning tasks where a large number of parameters need to be updated based on their gradients. Additional efforts need to made to incorporate NeuralSim with deep learning frameworks. Although we are not able to compare our work with these differentiable physics simulators, these difficulties demonstrate that dynamics and parameter learning with differentiable physics simulators are currently underexplored in the literature. 

\textbf{Societal impact:} We introduce a framework for data-driven dynamics modelling which uses physics-based priors to improve generalization, sample efficiency, and interpretability. Data-driven dynamics modelling, in general, can have a profound effect in learning-based control synthesis, especially in robotics and automation. However, our proposed framework is still a conceptual proposal and has a very low (around 2) Technology Readiness Level (TRL) \citep{hirshorn2016final}. We are yet to fully understand its limitations and failure scenarios that can significantly influence its real-world adoption. 
% The framework will have to go through several rounds of validation, before it can deployed in real-world systems.
%
%
%

\begin{ack}
The authors would like to thank Siemens Corporation, Technology for supporting this work. Funding in direct support of this work are from Siemens Corporation, Technology. There are no competing interests.
\end{ack}
\small
\setlength{\bibsep}{2pt plus 0.3ex}
\bibliography{ref}
\bibliographystyle{unsrtnat}

\normalsize
%\input{paper_checklist}

% \onecolumn
\newpage
\appendix
\appendixpage
\setcounter{equation}{0}
\renewcommand{\theequation}{S.\arabic{equation}}
\setcounter{figure}{0}
\renewcommand{\thefigure}{S.\arabic{figure}}
\setcounter{table}{0}
\renewcommand{\thetable}{S.\arabic{table}}
\setcounter{lstlisting}{0}
\renewcommand{\thelstlisting}{S.\arabic{lstlisting}}
\section{Notation}
\label{sec:notation}
\begin{tabular}[t]{l @{\hspace{.5em}} l}% 
$\mathbf{M}(\mathbf{x})$ & Inertia matrix \\
$V(\mathbf{x})$ & Potential energy \\
$\pmb{\mu}$ & coefficients of friction \\
$\mathbf{e}_P$ & coefficients of restitution \\
$\mathbf{p}_s = (\mathbf{M}(\mathbf{x}), V(\mathbf{x}))$ & System properties \\
$\mathbf{p}_c = (\pmb{\mu}, \mathbf{e}_P)$ & Contact properties \\
% $\mathbf{g}(\mathbf{x},\mathbf{v} ; \mathbf{p}_s)$ & First-order system dynamics \\
\midrule
$\mathbf{f}_C $ & Contact Impulses \\
$\mathbf{f}_E $ & Equality Constraint Impulses \\
$\mathbf{J}_C (\mathbf{x})$ & Contact Jacobian \\
$\mathbf{J}_E (\mathbf{x})$ & Equality Constraint Jacobian \\
$\mathbf{v}^-$/ $\mathbf{v}^+$ & velocities (in Cartesian space) before/after a general impulse \\
$\mathbf{v}_C^-$/ $\mathbf{v}_C^+$ & velocities (in contact space) before/after a general impulse \\
$\mathbf{v}_C^{c-}$/ $\mathbf{v}_C^{c+}$ & velocities (in contact space) before/after the compression phase \\
$\mathbf{v}_C^{r+}$ & velocities (in contact space) after the restitution phase \\
\end{tabular}

\section{Functional form of the constrained Lagrangian and Hamiltonian dynamics}
In this section, we present the functional form of system dynamics that we use in the experiments. Instead of using generalized coordinates, we use Cartesian coordinates. This is because the inertia matrix under Cartesian coordinates are constant and independent of the coordinates, which makes the learning of inertia easier, as pointed out in \citet{finzi2020simplifying}.
\subsection{Equality constraint Jacobian}
\label{sec:equ-constr}
Holonomic constraints are equality constraints which can be collected into a column vector $\Phi (\mathbf{x}) \in \mathbb{R}^E$ with equality  $\Phi (\mathbf{x}) = \mathbf{0}$. Differentiating this constraint w.r.t. time, we have%\dz{yields}
\begin{equation}
    \label{eqn:equality-constraint}
    \dot{\Phi} = (D_{\mathbf{x}}\Phi) \dot{\mathbf{x}} = (D_{\mathbf{x}}\Phi) \mathbf{v} = \mathbf{J}_E(\mathbf{x}) \cdot \mathbf{v}= \mathbf{0},
\end{equation}
where we denote the equality constraint Jacobian $\mathbf{J}_E(\mathbf{x}) := D_{\mathbf{x}}\Phi \in \mathbb{R}^{E\times D}$. Eqn. \eqref{eqn:equality-constraint} implies that holonomic constraints require the velocity $\mathbf{v}$ to be always in the null space of equality constraint Jacobian $\mathbf{J}_E(\mathbf{x})$. We will use this property to derive impulses caused by equality constraints. 
\subsection{Constrained Lagrangian dynamics}
The first-order dynamics can be obtained from  \citet{finzi2020simplifying}, which is
\begin{equation}
    %\dot{\textbf{z}} = \Psi(\textbf{z}) \Rightarrow
    \label{eqn:lag-constraint}
    \begin{pmatrix}
        \dot{\mathbf{x}} \\
        \dot{\mathbf{v}}
    \end{pmatrix} 
    = \mathbf{g}(\mathbf{x}, \mathbf{v}; \mathbf{p}_s) \!=\! 
    \begin{pmatrix}
        \mathbf{v} \\
        \mathbf{M}^{-1} \mathbf{J}_E^T [\mathbf{J}_E \mathbf{M}^{-1} \mathbf{J}_E^T ] ^{-1}
    [\mathbf{J}_E \mathbf{M}^{-1} \nabla_{\mathbf{x}} V \!-\! (D_{\mathbf{x}}(\mathbf{J}_E \cdot \mathbf{v})) \cdot \mathbf{v}] \!-\! \mathbf{M}^{-1}  \nabla_{\mathbf{x}} V
    \end{pmatrix}
\end{equation}

\subsection{Constrained Hamiltonian dynamics}
The Hamiltonian dynamics deal with position $\mathbf{x} \in \mathbb{R}^D$ and momentum $\mathbf{p}_{\mathbf{x}} = \mathbf{M} \mathbf{v}$ instead of $(\mathbf{x}, \mathbf{v})$. 
The derivation is not as straightforward as in the Lagrangian case. We denote $\mathbf{z} = (\mathbf{x}, \mathbf{p}_{\mathbf{x}})$. 
The Hamiltonian equals the total energy of the system and can be written as 
\begin{equation}
\label{eq:Lag-cart}
    H(\mathbf{x}, \mathbf{p}_{\mathbf{x}}) = \frac{1}{2} \mathbf{p}_{\mathbf{x}}^T \mathbf{M}^{-1}\mathbf{p}_{\mathbf{x}} + V(\mathbf{x}),
\end{equation}

For the $E$ holonomic constraints $\Phi (\mathbf{x}) = \mathbf{0}$, we can get another $E$ constraints on position and momentum, i.e., $\dot{\Phi}(\mathbf{x}, \mathbf{p}_{\mathbf{x}}) = 0$, and collect these $2E$ constraints in a vector $\Psi (\mathbf{z}) = (\Phi, \dot{\Phi})$.  Then the Hamiltonian dynamics in z can be written as the following differential equations
\begin{equation}
    \label{eq:ham-con}
    \dot{\mathbf{z}} = \mathbf{J} \nabla_{\mathbf{z}} H - \mathbf{J}(D_{\mathbf{z}}\Psi)^T [(D_{\mathbf{z}}\Psi) \mathbf{J}(D_{\mathbf{z}}\Psi)^T ]^{-1}(D_{\mathbf{z}}\Psi) \mathbf{J} \nabla_{\mathbf{z}} H,
\end{equation}
where $\mathbf{J}$ is a symplectic matrix
\begin{equation}
    \mathbf{J} = 
    \begin{bmatrix}
        \mathbf{0} &\mathbf{I}_D \\
        -\mathbf{I}_D & \mathbf{0}
    \end{bmatrix}
    ,
\end{equation}
and $\mathbf{I}_D$ is the $D \times D$ identity matrix. In order to convert the ODE into a set of ODE in $(\mathbf{x}, \mathbf{v})$, we introduce the matrix 
\begin{equation}
    \Tilde{\mathbf{M}}^{-1} = 
    \begin{bmatrix}
        \mathbf{I}_D & \mathbf{0} \\
        \mathbf{0} & \mathbf{M}^{-1} 
    \end{bmatrix}
    ,
\end{equation}
then we obtain the first order ODE 
\begin{equation}
    \begin{pmatrix}
        \dot{\mathbf{x}} \\
        \dot{\mathbf{v}}
    \end{pmatrix} =
    \begin{bmatrix}
        \mathbf{I}_D & \mathbf{0} \\
        \mathbf{0} & \mathbf{M}^{-1} 
    \end{bmatrix} 
    \begin{pmatrix}
        \dot{\mathbf{x}} \\
        \dot{\mathbf{p}_{\mathbf{x}}}
    \end{pmatrix}    
    =
    \Tilde{\mathbf{M}}^{-1} \mathbf{J} \nabla_{\mathbf{z}} H - \Tilde{\mathbf{M}}^{-1} \mathbf{J}(D_{\mathbf{z}}\Psi)^T [(D_{\mathbf{z}}\Psi) \mathbf{J}(D_{\mathbf{z}}\Psi)^T ]^{-1}(D_{\mathbf{z}}\Psi) \mathbf{J} \nabla_{\mathbf{z}} H
\end{equation}

\section{Mathematical Derivation of the differentiable contact model}
For simplicity, we present the model by referring to $\mathbf{x}$ and $\mathbf{v}$ as position and velocity in Cartesian space, but the derivation is valid for any other choice of coordinate system. A summary of the notation used here can be found in Section \ref{sec:notation}.

\subsection{Frictional contact and contact Jacobian}
Contacts in general can be expressed as inequalities $\Phi_C(\mathbf{x}) \geq \mathbf{0}$. A ball bouncing on the ground, for example, requires the whole ball to be above the ground. When the equality holds for a contact, we refer to the contact as an active contact, otherwise, an inactive contact. If there exists active contacts, contact impulses will cause an instantaneous velocity change. In practice, the set of active contacts is calculated by a collision detection (CD) module. 

A conceptual contact can contribute to one or more dimensions in the contact space, corresponding to one or more dimensions of contact impulse. Take Fig. \ref{fig:ball_pend} as an example. Mass 2 at the end of the pendulum would experience a contact impulse $\mathbf{f}_C=(f_n, f_t)$ in the two dimensional contact space - $f_n$ is the component normal to the contact surface, and $f_t$ is the friction impulse tangential to the contact surface. For 3D systems, the contact space is three dimensional with two tangential components. Assume that the contact space for all active contacts is $C$ dimensional, then we define contact Jacobian $\mathbf{J}_C(\mathbf{x}) \in \mathbb{R}^{C\times D}$, which maps velocities $\mathbf{v}$ in the coordinate space to $\mathbf{v}_C$ in the contact space,
\begin{equation}
    \mathbf{v}_C = \mathbf{J}_C(\mathbf{x}) \cdot \mathbf{v}.
\end{equation}
% we use the convention that the normal direction is from the first object to the second object.

\subsection{Project velocity change into contact space}
\label{sec:proj}
When there are active contacts, we construct the contact Jacobian $\mathbf{J}_C$ for active contacts. For brevity of notation, we drop explicit dependence on $\mathbf{x}$ from now onward. From Newton's second law, the change of momentum during contact equals the impulses, which can be described as
% The instantaneous velocity change 
% then the velocity jump during compression phase and restitution phase can both be described by the following equation. Todo: newton's second law. 
\begin{equation}
    \label{eqn:newton-2nd-law}
    \mathbf{M} \mathbf{v}^+ = \mathbf{M}\mathbf{v}^- + \mathbf{J}^T_C \mathbf{f}_C + \mathbf{J}^T_E \mathbf{f}_E,
\end{equation}
where $\mathbf{v}^-$ and $\mathbf{v}^+$ denote the Cartesian space velocity before and after the instantaneous velocity change, $\mathbf{M}$ is the inertia matrix, $\mathbf{J}^T_C$ maps contact impulses in the contact space $\mathbf{f}_C$ to contact impulses in Cartesian space and $\mathbf{J}^T_E$ maps equality constraint impulses $\mathbf{f}_E$ to equality constraint impulses in Cartesian space. 
The impulses $\mathbf{f}_C$ and $\mathbf{f}_E$ should not be confused with forces. An impulse is an integral of force over time, which contributes to the change in momentum. 

\begin{figure}[htbp]
    \centering
    \includegraphics[width=0.55\textwidth]{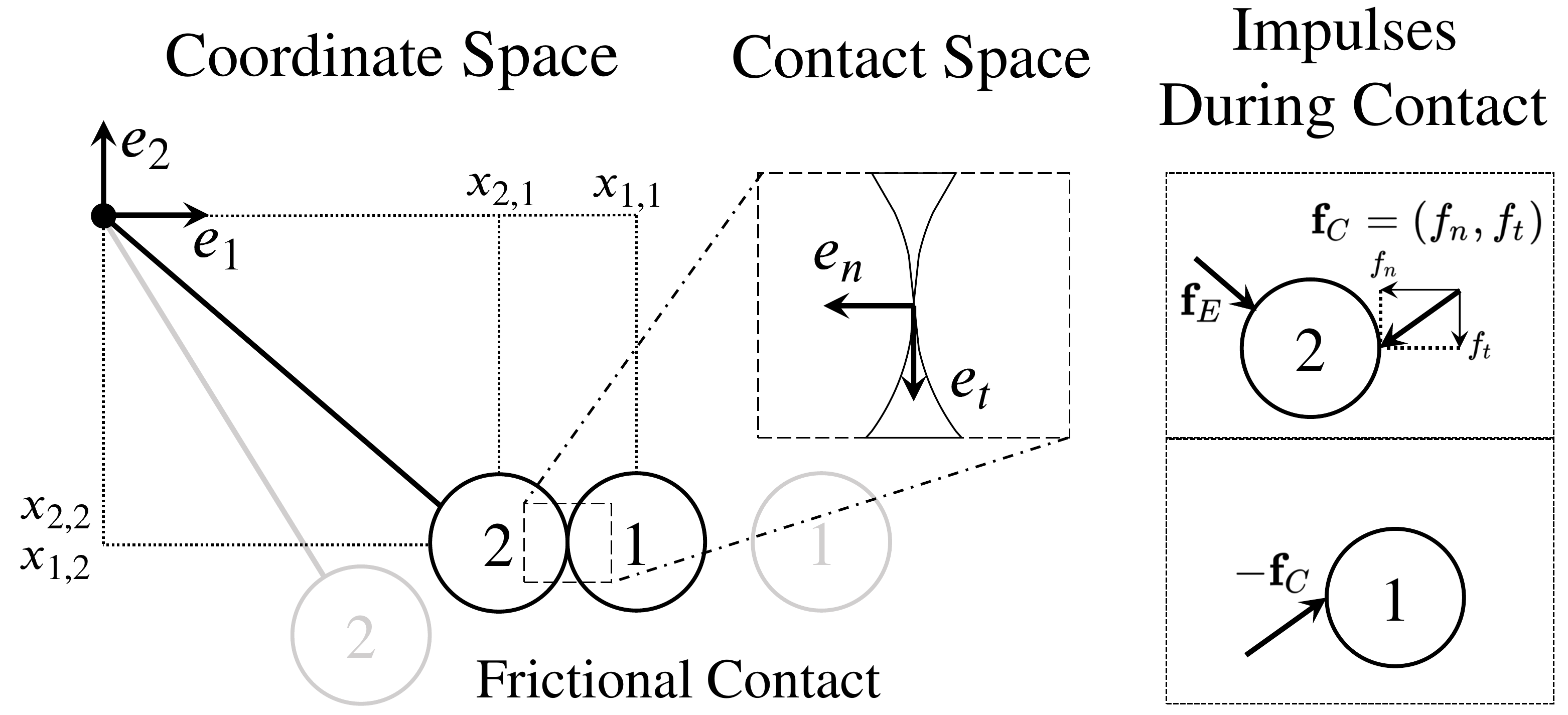}
    \caption{A ball collide with a pendulum. The equality constraint impulse $\mathbf{f}_E$ ensures that equality constraints are always satisfied.}
    \label{fig:ball_pend}
\end{figure}

The equality constraint impulses $\mathbf{f}_E$ is caused by contact impulse $\mathbf{f}_C$. See Fig. \ref{fig:ball_pend} for an intuitive example. Their dependence can be revealed from the fact that the velocity in Cartesian space at any time is in the null space of $\mathbf{J}_E$ (Sec. \ref{sec:equ-constr}.) We can left multiply the above equation by $\mathbf{J}_E \mathbf{M}^{-1}$ and solve for $\mathbf{f}_E$,
\begin{equation}
    \label{eqn:f_E}
    \mathbf{f}_E = - (\mathbf{J}_E \mathbf{M}^{-1} \mathbf{J}^T_E)^{-1} \mathbf{J}_E \mathbf{M}^{-1} \mathbf{J}^T_C \mathbf{f}_C.
\end{equation}
Thus, from Eqn. \eqref{eqn:newton-2nd-law} and \eqref{eqn:f_E}, we can express instantaneous velocity change as
\begin{equation}
    \label{eqn:v_change}
    \mathbf{v}^+ = \mathbf{v}^- + \mathbf{\hat{M
    }}^{-1} \mathbf{J}^T_C \mathbf{f}_C,
\end{equation}
where
\begin{equation}
    \mathbf{\hat{M}}^{-1} = \mathbf{M}^{-1} - \mathbf{M}^{-1} \mathbf{J}_E^T (\mathbf{J}_E \mathbf{M}^{-1} \mathbf{J}_E^T)^{-1} \mathbf{J}_E \mathbf{M}^{-1}.
\end{equation}
$\mathbf{\hat{M}}$ can be interpreted as the inertia that incorporates equality constraints. 

In order to solve for contact impulses, we left multiply Eqn. \eqref{eqn:v_change} by $\mathbf{J}_C$ to project the instantaneous velocity change into the contact space:
\begin{equation}
    \label{eqn:v-jump}
    \mathbf{v}_C^+ = \mathbf{v}_C^- + \mathbf{A} \mathbf{f}_C,
\end{equation}
where $\mathbf{A} = \mathbf{J}_C \mathbf{\hat{M
}}^{-1} \mathbf{J}^T_C$, which can be interpreted as the inverse inertia in the contact space. Our contact model solves contact impulses in two phases - the compression phase and the restitution phase, both of which can be described by Eqn. \eqref{eqn:v-jump}. Here we express the instantaneous velocity change in two phases as follows
\begin{align}
    \label{eqn:v-jump-comp}
    \mathbf{v}_C^{c+} &= \mathbf{v}_C^{c-} + \mathbf{A} \mathbf{f}_C^c, \\
    \label{eqn:v-jump-rest}
    \mathbf{v}_C^{r+} &= \mathbf{v}_C^{c+} - \mathbf{v}_C^* + \mathbf{A} \mathbf{f}_C^r,
\end{align}
where $\mathbf{f}_C^c$ and $\mathbf{f}_C^r$ are the contact impulses during the compression phase and the restitution phase and need to be solved by the contact model. The target velocity $\mathbf{v}_C^*$ is included in the restitution phase to compensate existing penetration in the simulation. See Appendix \ref{sec:pen_comp} for details on compensating penetration.
% Please see Supplementary Material for how this term influences simulation. 

\subsection{Contact model in compression phase}
From the maximum dissipation principle, the objective is to minimize the kinetic energy, which leads to the following optimization problem\footnote{Strictly speaking, this form is not correct because $\mathbf{A}$ is invertible only if there exists no equality constraint in the system. When equality constraints do exist, a pseudo-inverse of $\mathbf{A}$ should be used here, and the form Eqn. (5) can still be derived.}
% \begin{equation}
%     E = \frac{1}{2} (\mathbf{v}_C^{c})^T \mathbf{A}^{-1} \mathbf{v}_C^{c+} = \frac{1}{2} (\mathbf{f}_C^{c})^T \mathbf{A} \mathbf{f}_C^{c} + (\mathbf{f}_C^{c})^T \mathbf{v}_C^{c-}.
% \end{equation}
% Thus, we set up the 
\begin{align}
    \label{eqn:comp}
    &\underset{\mathbf{f}_C^c, \mathbf{v}_C^{c+}}{\textrm{Minimize }} \frac{1}{2} (\mathbf{v}_C^{c+})^T \mathbf{A}^{-1} \mathbf{v}_C^{c+} \\
    &\textrm{subject to } \eqref{eqn:v-jump-comp}, (3), (4). \nonumber
\end{align}

By substitute Eqn. \eqref{eqn:v-jump-comp} into \eqref{eqn:comp}, we get the optimization problem (5) in the paper. Similarly, optimization problem (7) can be derived. 

\section{Elasticity and coefficient of restitution}
The elasticity of a collision can be captured by the coefficient of restitution (COR). According to Newton's hypothesis \cite{newton1999principia}, COR is defined as the ratio of the normal relative velocity after the collision to that before the collision, ranging from 0 to 1. This definition of COR can cause unrealistic energy increases when the contact is frictional and the COR is close to 1 \cite{kane1985dynamics, stronge1991friction}. Alternatively, Poisson  \cite{poisson1817mechanics} divides the collision into two phases. The former, referred to as the compression phase, start with the first contact of the bodies and stops at the greatest compression. The latter, referred to as the restitution phase, start right after the compression phase till the separation of bodies. According to Poisson's hypothesis, the COR is defined as the ratio of the normal contact impulse in the restitution phase to that in the compression phase. Poisson's hypothesis is favored in simulation since it will not lead to unrealistic energy increase. For a detailed comparison of different hypotheses, please refer to \cite{djerassi2009collisionA, djerassi2009collisionB}. In this paper, we define COR $e_P$ in accordance with Poisson's hypothesis. 

\section{Proof of positive semi-definiteness of $\mathbf{A}$}
\label{sec:proof}
By definition, we have $\mathbf{A} = \mathbf{J}_C \mathbf{\hat{M
    }}^{-1} \mathbf{J}^T_C \in \mathbb{R}^{C\times C}$, where
\begin{equation}
    \mathbf{\hat{M}}^{-1} = \mathbf{M}^{-1} - \mathbf{M}^{-1} \mathbf{J}_E^T (\mathbf{J}_E \mathbf{M}^{-1} \mathbf{J}_E^T)^{-1} \mathbf{J}_E \mathbf{M}^{-1}.
\end{equation}
For any real physical system, the inertia matrix $\mathbf{M}$ is symmetric and positive definite. Thus, its inverse exists and can be decomposed using Cholesky decomposition $\mathbf{M} = \mathbf{L} \mathbf{L}^T$. We can then express the inverse inertia that incorporates equality constraints as $\mathbf{\hat{M}}^{-1} = \mathbf{L} (\mathbf{I} - \mathbf{P}) \mathbf{L}^T$, where $\mathbf{P}$ is a projection matrix
\begin{equation}
    \mathbf{P} = \mathbf{L}^T \mathbf{J}_E^T (\mathbf{J}_E \mathbf{M}^{-1} \mathbf{J}_E^T)^{-1} \mathbf{J}_E \mathbf{L},
\end{equation}
which satisfies $\mathbf{P}^2 = \mathbf{P}$. A property of projection matrices is that the eigenvalues can only take two values: 1 or 0. By eigen-decomposition, $\mathbf{P}$ and $\mathbf{I} - \mathbf{P}$ can be written as 
\begin{equation}
    \mathbf{P} = 
    \begin{pmatrix}
        \mathbf{V}_0 & \mathbf{V}_1
    \end{pmatrix}
    \begin{pmatrix}
        \mathbf{0}& \mathbf{0} \\
        \mathbf{0}& \mathbf{I}
    \end{pmatrix}
    \begin{pmatrix}
        \mathbf{V}_0^T \\ \mathbf{V}_1^T
    \end{pmatrix}
    = 
    \mathbf{V}_1 \mathbf{V}_1^T,
\end{equation}
\begin{equation}
    \mathbf{I} - \mathbf{P} = 
    \begin{pmatrix}
        \mathbf{V}_0 & \mathbf{V}_1
    \end{pmatrix}
    \begin{pmatrix}
        \mathbf{I}& \mathbf{0} \\
        \mathbf{0}& \mathbf{I}
    \end{pmatrix}
    \begin{pmatrix}
        \mathbf{V}_0^T \\ \mathbf{V}_1^T
    \end{pmatrix}
    -
    \begin{pmatrix}
        \mathbf{V}_0 & \mathbf{V}_1
    \end{pmatrix}
    \begin{pmatrix}
        \mathbf{0}& \mathbf{0} \\
        \mathbf{0}& \mathbf{I}
    \end{pmatrix}
    \begin{pmatrix}
        \mathbf{V}_0^T \\ \mathbf{V}_1^T
    \end{pmatrix}
    = \mathbf{V}_0 \mathbf{V}_0^T,
\end{equation}
where $\mathbf{V}_0 \in \mathbb{R}^{D\times (D-E)}$. So we can decompose $\mathbf{A}$ into $\mathbf{A} = \mathbf{A}_d^T \mathbf{A}_d$, where $\mathbf{A}_d = \mathbf{V}_0^T \mathbf{L}^T \mathbf{J}^T_C$.
Then for any vector $\mathbf{c} \in \mathbb{R}^C$, we have 
\begin{equation}
    \mathbf{c}^T \mathbf{A} \mathbf{c} = (\mathbf{A}_d \mathbf{c})^T \mathbf{A}_d \mathbf{c} \geq 0,
\end{equation}
which proves that $\mathbf{A}$ is positive semi-definite. If the system does not have equality constraints, $\mathbf{A}$ has full rank and can be decomposed using Cholesky decomposition. 

Note that in this forward pass, we need to use \Verb+torch.symeig+ on matrix $\mathbf{P}$. However, 
\Verb+torch.symeig+ operation does not support backward gradient calculation with non-distinct eigenvalues. In practice, we use the implementation in \cite{kasim2020derivatives} to calculate the gradient of \Verb+torch.symeig+ operation.

\section{Solving contact impulses using CvxpyLayers}
In this section, we show an implementation of setting up the differentiable optimization problem in the compression phase using CvxpyLayers and PyTorch. We then show how we use this implementation in CM and CMr.
{
\small
\begin{lstlisting}[language=python, caption=Implementation of solving compression phase impulse using CvxpyLayers]
import torch
import cvxpy as cp
from cvxpylayers.torch import CvxpyLayer

def solve_compression_impulse(
        A_d: torch.Tensor, # shape (D-E, C) or (C, C), decomposition of matrix A
        v_: torch.Tensor, # shape (C, 1), velocity before impulse in contact space
        mu: torch.Tensor, # shape (n_cld, 1), coefficient of friction
        n_cld: int, # number of active (conceptual) contacts
        d: int, # dimension of each conceptual contact space, can take value 2 or 3 
    ):
    C = v_.shape[0] # C = n_cld*d
    f = cp.Variable((C, 1)) # impulse variable to be solved
    A_d_p = cp.Parameter(A_d.shape) 
    v_p = cp.Parameter((C, 1))
    mu_p = cp.Parameter((mu.shape[0], 1)) 
    # set up objective, constraints, cvx problem and cvxpylayer
    objective = cp.Minimize(0.5 * cp.sum_squares(A_d_p @ f) + cp.sum(cp.multiply(f, v_p)))
    constraints = [cp.SOC(cp.multiply(mu_p[i], f[i*d]), f[i*d+1:i*d+d]) 
                    for i in range(n_cld)] \
                  + [f[i*d] >= 0 for i in range(n_cld)]
    problem = cp.Problem(objective, constraints)
    cvxpylayer = CvxpyLayer(problem, parameters=[A_d_p, v_p, mu_p], variables=[f])
    # forward pass
    impulse, = cvxpylayer(A_d, v_, mu)
    return impulse
\end{lstlisting}

\begin{lstlisting}[language=python, caption=pseudocode of solving compression phase impulse in CM]
...
# get A_d as stated in Section E.
A_d  # shape (D-E, C)
impulse = solve_compression_impulse(A_d, v_, mu, n_cld, d)
\end{lstlisting}

\begin{lstlisting}[language=python, caption=pseudocode of solving compression phase impulse in CMr]
...
# regularization
R = torch.eye(A.shape[0]).type_as(A)*1e-2 # shape (C, C)
A_d = torch.cholesky(A+R, upper=True) # shape (C, C)
impulse = solve_compression_impulse(A_d, v_, mu, n_cld, d)
\end{lstlisting}
}
The only difference between CM and CMr is how we construct the matrix $\mathbf{A}_d$. In CM, $\mathbf{A}_d$ is constructed as described in Section \ref{sec:proof}, while in CMr, $\mathbf{A}_d$ is constructed by adding a regularization and performing Cholesky decomposition.

\section{Penetration compensation}
\label{sec:pen_comp}
To compensate for an existing penetration during restitution phase in the simulation, we use the target velocity $\mathbf{v}_C^* \in \mathbb{R}^C$ and the optimization problem (7). In this section, we discuss how to calculate the target velocity $\mathbf{v}_C^*$ so that it does not violate the equality constraints of the system. The calculation of $\mathbf{v}_C^*$ might be nontrivial; however, in the backward pass, the gradients of $\mathbf{v}_C^*$ are not required for learning contact properties. Thus, in practice, we do not calculate the backward gradients for every calculation introduced in this section. 
%We compensate existing penetration in simulation during the restitution phase. This is achieved using the target velocity $\mathbf{v}_C^* \in \mathbb{R}^C$ and the optimization problem (17). In this section, we discuss how to choose the target velocity $\mathbf{v}_C^*$ that does not violate the equality constraints of the system. The calculation of $\mathbf{v}_C^*$ might be nontrivial, but in the backward pass, the gradients of $\mathbf{v}_C^*$ are not required for learning contact properties. Thus, in practice, we do not calculate the backward gradients for all calculations introduced in this section. 

% > \mathbf{0}
To choose the target velocity $\mathbf{v}_C^*$, we first come up with a desired velocity $\mathbf{v}_C^d \in \mathbb{R}^C$. For each direction normal to contact surfaces, the component in $\mathbf{v}_C^d$ is calculated as the depth of penetration divided by integration time interval. For each tangential dimension, the component in $\mathbf{v}_C^d$ is set to zero. This choice of $\mathbf{v}_C^d$ will fix penetration in the next time step. The downside is that for totally inelastic contacts, in the next few time steps, the bodies in collision might separate (because the relative velocity normal to the contact surface is greater than zero), which make the contact looks like partially elastic. This phenomenon can be avoided by using more than one time step to compensate the penetration, i.e., by setting the components in $\mathbf{v}_C^d$ to be a fraction of the depth of the penetration, as shown in the figure below. 
\begin{figure}[h]
    \centering
    \includegraphics[width=0.91\textwidth]{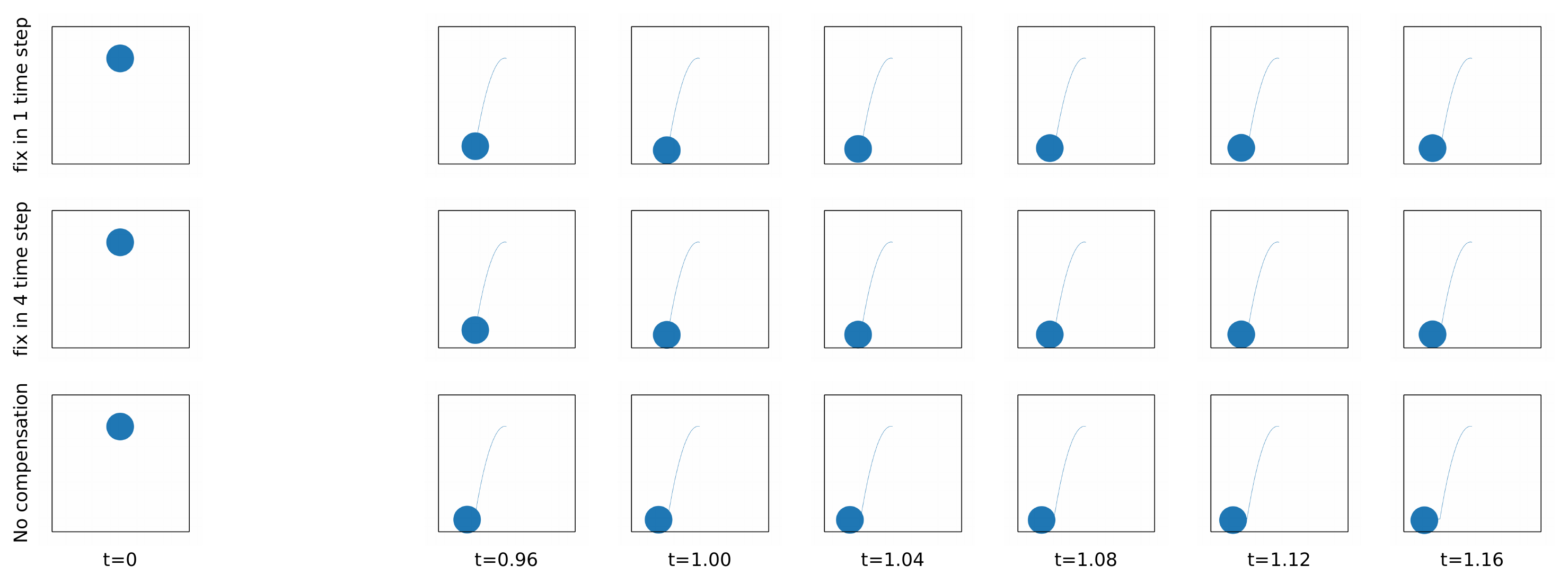}
    \caption{Different $\mathbf{v}_C^d$ for compensation in a bouncing point mass with gravity and COR=0. \textbf{First row}: fixing penetration in 1 time step. The circle bounces off the ground after touching the ground. \textbf{Second row}: fixing penetration in 4 time steps. The penetration is fixed and the point mass doesn't bounce up.  \textbf{Third row}: no penetration compensation. The penetration are not fixed over time. }
    \label{fig:compensation}
\end{figure}
The reason that we cannot use $\mathbf{v}_C^d$ as the target velocity is that $\mathbf{v}_C^d$ might violate equality constraints. Take a ball collide with a pendulum as an example (Fig. 1 in main paper), $\mathbf{v}_C^d$ would violate the equality constraint of the pendulum, i.e., the velocity of object 2 can only be perpendicular to the pendulum. Thus, we need to transform $\mathbf{v}_C^d$ into target velocity $\mathbf{v}_C^*$ that satisfies the equality constraints. 

% The calculations are different for $C<=D$ ($\mathbf{J}_C$ as a fat matrix) and $C>D$ ($\mathbf{J}_C$ as a tall matrix), which are presented respectively in Section \ref{sec:cal-C-D} and Section \ref{sec:cal-D-C}. 

The idea of obtaining a target velocity $\mathbf{v}_C^*$ is to project the desired velocity $\mathbf{v}_C^d$ into Cartesian space, make corrections to satisfy equality constraints and then project it back to the contact space. In Section \ref{sec:proj}, we showed how to project impulses between contact space and Cartesian space. However, strictly speaking, the projection defined would introduce a scaling if one, say, project a vector from contact space to Cartesian space and back to contact space. This is not a problem for solving contact impulses, but it will be problematic if we have this scaling in calculating target velocity $\mathbf{v}_C^*$. To fix this issue, we need to introduce the pseudoinverse of $\mathbf{J}_C$. Let's assume we have some form of pseudoinverse $\mathbf{J}_C^+$ (we will define it later.)
Then the target velocity in Cartesian space is the sum of the desired velocity $\mathbf{v}_C^d$ projected into Cartesian space and a correction term. 
\begin{equation}
    \mathbf{v}^* = \mathbf{J}_C^+  \mathbf{v}_C^d + \mathbf{J}_E^T \mathbf{v}_E^d
\end{equation}
The equality constraints require $\mathbf{J}_E \mathbf{v}^* = \mathbf{0}$, from which we can solve for $\mathbf{v}_E^d = - (\mathbf{J}_E \mathbf{J}_E^T)^{-1} \mathbf{J}_E \mathbf{J}_C^+ \mathbf{v}_C^d$, then we project the target velocity in Cartesian space $\mathbf{v}^*$ into contact space and get
\begin{equation}
    \mathbf{v}_C^* = \mathbf{J}_C \mathbf{v}^* = \mathbf{J}_C (\mathbf{I} - \mathbf{J}_C (\mathbf{J}_E \mathbf{J}_E^T)^{-1} \mathbf{J}_E) \mathbf{J}_C^+ \mathbf{v}_C^d
\end{equation}

The form of the pseudoinverse $\mathbf{J}_C^+$ we use here is dependent on the shape of $\mathbf{J}_C$. 
We define the pseudoinverse of $\mathbf{J}_C$ as 
\begin{equation}
    \mathbf{J}_C^+ = 
    \begin{cases}
    \mathbf{J}_C^T (\mathbf{J}_C \mathbf{J}_C^T)^{-1}, \text{ if } C\leq D \\
    (\mathbf{J}_C^T \mathbf{J}_C)^{-1} \mathbf{J}_C^T, \text{ if } C > D
    \end{cases}
\end{equation}

When $C\leq D$, the dimension of contact space is smaller than that of Cartesian space, we can verify that projecting a velocity from contact space to Cartesian space and back to contact space equals the original velocity, i.e., $\mathbf{J}_C \mathbf{J}_C^+ = \mathbf{I}_C$. When $C > D$, the dimension of contact space is greater than that of Cartesian space, we can verify that projecting a velocity from Cartesian space to contact space and back to Cartesian space equals the original velocity, i.e., $\mathbf{J}_C^+ \mathbf{J}_C = \mathbf{I}_D$.

\section{Simulated systems}
\textbf{Bouncing point masses.} This system is often referred to as bouncing balls in previous works \cite{battaglia2016interaction, chang2016compositional}. We call it bouncing point masses instead since each object is essentially a circle with a point mass at the center and cannot rotate like real balls. For $n$ objects bouncing in the box, there exists $n(n-1)/2$ possible contacts between objects and $4n$ possible contacts between objects and walls. These contacts cannot be all active simultaneously. 
% For 5 objects, we estimate a maximum of 8 contacts to be simultaneously active. 
We set up two tasks of 5 bouncing point masses with different configurations, which will be referred to as BP5-e and BP5. BP5-e is a homogeneous setting, where the masses and radii are the same for all objects, and contact properties ($e_P=1$ and $\mu=0$) are the same for all contacts. This task conserves energy since no energy is lost during the collisions and during collision-free periods. BP5 is a heterogeneous setting where the masses and radii are different for different points and contact properties are different for different contacts. 
% The heterogeneity makes the dynamics more challenging to learn. 

\textbf{Bouncing disks.} A real 2D disk has mass spread over the circle and thus can rotate, especially when frictional contacts are involved. Thus we extend the bouncing point masses system to bouncing disks, where all the disks can rotate. We use the extended bodies representation introduced in \cite{finzi2020simplifying} to embed the motion of disks in Cartesian coordinates. The idea is to use the motion of 3 points - the center of mass as well as the unit vectors aligned with two principle axes - to describe the motion of a disk. Since the relative position of these 3 points are fixed, this representation will introduce 3 equality constraints for each disk. A contact impulse will be distributed properly into these 3 points in a way that obeys the law of physics. Please refer to \cite{finzi2020simplifying} for more details on this representation. We simulate 5 heterogeneous bouncing disks with heterogeneous contact properties. This task is referred to as BD5.

\textbf{Chained pendulums with ground.} The 2-pendulum colliding with ground has been used to study and analyze contact models more than three decades ago \cite{kane1985dynamics}. Until recently, some works \cite{finzi2020simplifying,zhong2020benchmarking} have studied learning dynamics of N-pendulums without contacts. Here we simulate a 3-pendulum system above the ground where the lowest pendulum can collide with the ground. The masses are located at the joints and the sizes of the joints are different. We follow the convention to assume that pendulums cannot collide with each other. We propose two tasks: CP3-e with $e_P=1$ and $\mu=0$, where energy is conserved, and CP3 with $e_P=0$ and $\mu=0.5$.
% We simulate a energy conserving configuration, referred to as CP3-E, where $e=1$ and $\mu=0$, as well as a configuration, referred to as CP3, with frictional inelastic collision ($e=0$ and $\mu=0.5$.)

\textbf{Gyroscope with a wall.} Gyroscope is a 3D system that exhibits complex dynamics such as precession and nutation. In order to test our contact model in 3D space, we extend the gyroscope system by putting a wall near it so that collisions can happen. 
The motion of the gyroscope is embedded in Cartesian coordinates using the extended bodies representation\cite{finzi2020simplifying}. This representation introduces 6 equality constraints. As the gyroscope is attached to a ball joint, one more equality constraint is introduced. We propose two tasks: Gyro-e with $e_P=1$ and $\mu=0$, which conserves energy, and Gyro with $e_P=0.8$ and $\mu=0.1$.

\textbf{Rope.} 
Our contact model can also capture limits in joint angles and distances, which we show in this rope system. The motion of the rope is described by 10 equally spaced points along the rope. The distance between adjacent points are not fixed as in the chained pendulums system. Instead, the rope can be stretched. The stretch is modelled by elastic springs connecting each pair of adjacent points. We set the maximum stretch and minimum stretch to be 1.2 and 0.8, respectively, which implies that two adjacent points are not allowed to be pushed or pulled by more than $20\%$ of their distance at rest. We also assume that two adjacent segments cannot be bent over a predefined angle (0.2rad). The above stretch and bending constraint can be handled by our contact model with $e=0$ and $\mu=0$. 
% This bending constraint can also be modelled by our contact model with a 2D contact space where $e=0$ and $\mu=0$. 
During simulation of the rope, a total of 19 ``contacts" can be active at the same time, which makes it nontrivial to solve for contact impulses. The force of the spring is modelled via the potential energy, which results in a potential energy function that is not linear in the location of points. This is the only system tested in this work that has a nonlinear potential energy function. 
% We will show we can learn the potential energy function and the contacts properties simultaneously using our proposed method. 
This setup is similar to the rope proposed in \cite{yang2020learning}, and differs from string proposed in \cite{battaglia2016interaction}, as the latter impose no bending constraint.

\section{Mass ratio details}
Here we show the learned mass ratios in BP5-e, BP5, CP3-e and CP3 tasks. We can see the learned mass ratios match the ground truth with high accuracy. This shows our framework learns interpretable mass ratios.

\begin{table}[h]
\centering
\caption{Learned mass ratios in BP5-e}
% \vspace{-1.7em}
\small
\begin{tabularx}{0.63\textwidth}{c | Y | Y | Y | Y}
    \toprule[1pt]
     \textbf{Mass ratio} &  $m_2/m_1$ & $m_3/m_1$ & $m_4/m_1$ & $m_5/m_1$ \\
     \midrule[0.5pt]
     True & 1.0000 & 1.0000 & 1.0000 & 1.0000 \\
     \midrule[0.5pt]
     CM-CD-CLNN & 1.0000 & 1.0000 & 1.0002 & 1.0003 \\
     CM-CD-CHNN & 0.9998 & 1.0000 & 1.0000 & 1.0000 \\
     CMr-CD-CLNN & 1.0000 & 0.9993 & 0.9991 & 0.9989 \\
     CMr-CD-CHNN & 1.0004 & 0.9994 & 0.9999 & 0.9997 \\
    \bottomrule[1pt]
\end{tabularx}
% \vspace{-2em}
\end{table}

\begin{table}[h]
\centering
\caption{Learned mass ratios in BP5}
% \vspace{-1.7em}
\small
\begin{tabularx}{0.63\textwidth}{c | Y | Y | Y | Y}
    \toprule[1pt]
     \textbf{Mass ratio} &  $m_2/m_1$ & $m_3/m_1$ & $m_4/m_1$ & $m_5/m_1$ \\
     \midrule[0.5pt]
     True & 2.0000 & 6.0000 & 8.0000 & 10.0000 \\
     \midrule[0.5pt]
     CM-CD-CLNN & 2.0000 & 6.0036 & 8.0014 & 10.0024 \\
     CM-CD-CHNN & 2.0005 & 6.0020 & 8.0015 & 10.0029 \\
     CMr-CD-CLNN & 1.9998 & 6.0004 & 8.0033 & 9.9997 \\
     CMr-CD-CHNN & 2.0002 & 6.0001 & 7.9985 & 10.0010 \\
    \bottomrule[1pt]
\end{tabularx}
% \vspace{-2em}
\end{table}
\begin{table}[h]
\centering
\caption{Learned mass ratios in CP3 and CP3-e}
% \vspace{-1.7em}
\small
\begin{tabularx}{0.63\textwidth}{c | Y | Y | Y | Y}
    \toprule[1pt]
     \textbf{Mass}& \multicolumn{2}{c|}{CP3} & \multicolumn{2}{c}{CP3-e} \\
     \textbf{ratio} &  $m_2/m_1$ & $m_3/m_1$ & $m_2/m_1$ & $m_3/m_1$ \\
     \midrule[0.5pt]
     True & 0.6500 & 0.7500 & 2.0000 & 1.5000 \\
     \midrule[0.5pt]
     CM-CD-CLNN & 0.6500 & 0.7502 & 2.0006 & 1.4990 \\
     CM-CD-CHNN  & 0.6499 & 0.7500 & 1.9996 & 1.4994 \\
     CMr-CD-CLNN & 0.6500 & 0.7521 & 2.0002 & 1.5001 \\
     CMr-CD-CHNN & 0.6503 & 0.7526 & 2.0009 & 1.5009\\
    \bottomrule[1pt]
\end{tabularx}
% \vspace{-2em}
\end{table}

\section{Analysis of LCP baseline}
We use the formulation and codebase provided in \cite{de2018end}. The core implementation of differentiable LCP solver is the \Verb+LCPFunction+ class, which is a subclass of \Verb+torch.autograd.Function+. The forward pass of the \Verb+LCPFunction+ solves a LCP problem and the backward pass computes the gradients. Both the forward pass and the backward pass leverage the primal dual interior point method (pdipm) to compute relevant quantities. However, the provided codebase is outdated and is not compatible with the latest Pytorch release. In order to leverage the codebase to compare it against our method, we first update the core implementation to make it compatible with the latest Pytorch release. We have done a sanity check on the examples provided in the codebase to make sure the updated forward pass and backward pass gives the same results as in the original codebase. 

We formulate our 2D and 3D contact problems as LCP problems and use the updated codebase for simulation. As the standard LCP formulation adopts Newton's hypothesis to model elasticity, we adopts Newton's hypothesis in our LCP formulation as well. We plan to use the bouncing point masses system to compare differentiable LCP and our method, since here Newton's hypothesis and Poisson's hypothesis results in the same contact impulses. (In a general system, such as the gyroscope with wall, these two hypotheses result in different contact impulses.)
We observe that our LCP formulation generate expected rigid body motions, which shows that the forward pass of \Verb+LCPFunction+ works well with the CLNN/CHNN dynamics and the extended bodies representation \cite{finzi2020simplifying}. However, when we try to learn system and contact properties from generated trajectories, we observe that the backward pass of \Verb+LCPFunction+ always gives gradients as NaNs. To be specific, the place where NaNs first show up is the \Verb+pdipm.solve_kkt()+ function call in the backward pass of \Verb+LCPFunction+. This indicates a problem with the computation of gradients in the LCP solver. Further investigation is required to see if this is a problem about the primal dual interior point method (pdipm) itself or numerical stability in the implementation.

\section{Robustness analysis details}
In this section, we show additional robustness results. These results shows that our model is robust under model mismatch (LCP generated training data) and noise.
\begin{table}[h]
\centering
\caption{Robustness on LCP data (contact properties)}
% \vspace{-1.3em}
% \hspace{-1.3em}
\small
\begin{tabularx}{0.87\textwidth}{c | Y | Y | Y | Y | Y | Y | Y | Y }
    \toprule[1pt]
     & \multicolumn{2}{c|}{CP3} & \multicolumn{2}{c|}{CP3-e} & \multicolumn{2}{c|}{BP5-e} &
     \multicolumn{2}{c}{Gyro} \\
     & $\mu$ & $e_P$ & $\mu$ & $e_P$ & $\mu$ & $e_P$ & $\mu$ & $e_P$  \\
     \midrule[0.5pt]
     True & 0.500 & 0.000 & 0.000 & 1.000 & 0.000 & 1.000 & 0.100 & 0.800\\
     \midrule[0.5pt]
     Trained by CM data & 0.500 & 0.004 & 0.000 & 1.000 & 0.000 & 1.000 & 0.100 & 0.800 \\
    Trained by LCP data & 0.500 & 0.005 & 0.003 & 1.000 & 0.000 & 1.000 & 0.100 & 0.822\\
    \bottomrule[1pt]
\end{tabularx}
% \vspace{-1.25em}
\end{table}
\begin{table}[h]
\centering
\caption{Robustness on LCP data (trajectory relative error w. 95\% conficence interval)}
% \vspace{-1.3em}
% \hspace{-1.3em}
\small
\begin{tabularx}{1.0\textwidth}{Y | c | c | c | c }
    \toprule[1pt]
    & CP3 & CP3-e & BP5-e &
     Gyro \\
     \midrule[0.2pt]
     Trained by CM data, validated on CM val. data  & 2.34e-5(1.29e-5) &	3.85e-3(9.51e-4) &	2.83e-3(3.64e-4) &	2.39e-3(1.36e-3) \\
     \midrule[0.2pt]
    Trained by LCP data, validated on CM val. data & 2.54e-3(3.73e-3) &	3.73e-3(6.92e-4) &	1.57e-2(5.00e-3) &	5.50e-3(1.67e-3) \\
    \midrule[0.2pt]
    Trained by LCP data, validated on LCP val. data & 7.35e-4(3.32e-4) &	1.31e-3(3.26e-4) &	5.46e-3(2.30e-3) &	4.51e-4(2.65e-4)\\
    \bottomrule[1pt]
\end{tabularx}
% \vspace{-1.25em}
\end{table}
\begin{table}[h]
\centering
\caption{Robustness on noisy data (contact properties)}
% \vspace{-1.7em}
\small
\begin{tabularx}{0.57\textwidth}{c | Y | Y | Y | Y}
    \toprule[1pt]
     \textbf{Noisy}& \multicolumn{2}{c|}{CP3} & \multicolumn{2}{c}{CP3-e} \\
     \textbf{Data} &  $\mu$ & $e_P$ & $\mu$ & $e_P$ \\
     \midrule[0.5pt]
     True & 0.500 & 0.000 & 0.000 & 1.000 \\
     \midrule[0.5pt]
     0 & 0.500 & 0.023 & 0.002 & 1.000 \\
     $\mathcal{N}(0, 0.01)$  & 0.496 & 0.036 & 0.004 & 1.000 \\
     $\mathcal{N}(0, 0.05)$   & 0.462 & 0.061 & 0.004 & 1.000 \\
    \bottomrule[1pt]
\end{tabularx}
% \vspace{-2em}
\end{table}
\begin{table}[h]
% \vspace{-1.5em}
\caption{Robustness on the Regularizer in CMr (contact properties)}
\centering
\small
\begin{tabularx}{0.57\textwidth}{c | Y | Y | Y | Y }
    \toprule[1pt]
     \textbf{Regularizer}& \multicolumn{2}{c|}{CP3} &
     \multicolumn{2}{c}{Gyro} \\
     \textbf{Ablation}& $\mu$ & $e_P$ & $\mu$ & $e_P$  \\
     \midrule[0.5pt]
     True & 0.500 & 0.000 & 0.100 & 0.800\\
     \midrule[0.5pt]
     $\epsilon=0.001$ & 0.500 & 0.007 & 0.100 & 0.811 \\
     $\epsilon=0.01$  & 0.500 & 0.023 & 0.099 & 0.892 \\
     $\epsilon=0.1$   & 0.501 & 0.180 & 0.100 & 0.886 \\
     learnable & 0.497 & 0.453 & 0.100 & 0.861\\
    \bottomrule[1pt]
\end{tabularx}
% \vspace{-2.25em}
\end{table}

\section{Scalability details}
In this section we show more results on scalability. We use the Rope system to explore how the trajectory relative error changes with the neural network size, number of training trajectory $N$ and the degrees of freedom $D$ (equivalently, the number of contacts). All results reported in the tables are averaged over 100 test trajectories with 95\% confidence interval.

Table~\ref{tab:scale-network} shows the effect of network size. Our default network used in CLNN (to approximate potential energy) is an MLP with 3 layers with hidden sizes of 256. We enlarge the network with 6 layers with hidden sizes of 512. We find that large networks doesn’t improve our models performance. This is likely because with the default size, the potential energy is already estimated well enough. 

Table~\ref{tab:scale-traj} shows how the trajectory relative error varies with different number of training trajectories. As expected, for all of the four models, the error decreases with increasing number of training trajectories. For CM-CD-CLNN, it seems the decreasing trend hasn't converge yet. For MLP-CD-CLNN, the errors doesn't change much from N=800 to N=12800. For IN-CP-CLNN and IN-SP-CP, it is hard to tell if the decrease has converged or not, but it is clear that they have the highest errors across four models (each row). If we compare CM-CD-CLNN and MLP-CD-CLNN, we can clearly see that the gap between our model and the baseline decreases from N=25 to N=800 and increases from N=800 to N=12800. If we compare our model and the other two baselines, the gap decreases from N=25 to N=800, but the trend from N=800 to N=12800 is unclear. The decreasing trend might be unexpected. The underlying reason is that our model performs well with a small amount data. For our model, the difference between N=25 and N=12800 is just 2.9e-4. For baseline models, this difference (between N=25 and N=12800) is at least one order of magnitude higher. Since our model has strong physics priors, this result is expected and shows that our model is data efficient. 

Table~\ref{tab:scale-contacts} shows the trajectory relative error (the same metric used in Figure 3) of ropes discretized in different number of segments (The configurations in Table 3). We find that the performance gap between our model and baselines is the smallest in D=400 scenario. This indicates that our model might not have a clear advantage over baselines in contact-rich scenarios. 

These set of results shows that our model might not perform well for contact-rich scenarios. We’d also like to point out that even if our model does not have a clear advantage in contact-rich scenarios, our main contribution is to demonstrate the framework’s ability in simultaneously learning of unknown system dynamics and contact properties from trajectory data.

\begin{table}[h]
% \vspace{-1.5em}
\caption{Scalability - large networks}
\label{tab:scale-network}
\centering
\small
\begin{tabularx}{0.57\textwidth}{c | Y }
    \toprule[1pt]
     & CM-CD-CLNN  \\
     \midrule[0.5pt]
     default network & 3.20e-3(5.69e-4) \\
     large size  & 6.89e-3(5.07e-4) \\
    \bottomrule[1pt]
\end{tabularx}
% \vspace{-2.25em}
\end{table}

\begin{table}[h]
% \vspace{-1.5em}
\caption{Scalability - different training trajectories}
\label{tab:scale-traj}
\centering
\small
\begin{tabularx}{0.90\textwidth}{c | Y | Y | Y | Y }
    \toprule[1pt]
     & CM-CD-CLNN & MLP-CD-CLNN & IN-CP-CLNN & IN-SP-CP  \\
     \midrule[0.5pt]
     N=25 & 1.91e-3(3.06e-4) &	2.96e-2(8.04e-3) &	3.73e-2(3.41e-3) &	5.96e-2(7.10e-3) \\
     N=50  & 1.87e-3(9.60e-4) &	1.25e-2(2.32e-3) &	3.06e-2(2.35e-3) &	6.49e-2(7.71e-3) \\
     N=100   & 2.04e-3(2.91e-4) &	7.36e-3(9.63-e4) &	2.10e-2(1.13e-3) &	2.82e-2(1.92e-3) \\
     N=200 & 1.92e-3(3.12e-4) &	8.66e-3(1.04e-3) &	2.68e-2(1.61e-3) &	3.59e-2(2.39e-3)\\
     N=400 & 1.90e-3(3.40e-4) &	7.02e-3(7.76e-4) &	1.44e-2(1.01e-3) &	2.71e-2(1.61e-3) \\
     N=800 & 3.20e-3(5.69e-4) &	5.98e-3(5.21e-4) &	6.87e-3(5.11e-4) &	8.94e-3(5.30e-4) \\
     N=1600 & 2.27e-3(3.51e-4) &	5.94e-3(5.26e-4) &	6.86e-3(4.88e-4) &	9.19e-3(5.27e-4) \\
     N=3200 & 1.64e-3(2.73e-4) &	5.98e-3(5.27e-4) &	6.21e-3(5.14e-4) &	7.33e-3(5.26e-4) \\
     N=6400 & 1.57e-3(2.27e-4) &	5.92e-3(5.26e-4) &	5.90e-3(5.27e-4) &	7.72e-3(5.85e-4) \\
     N=12800 & 1.20e-3(2.88e-4) &	5.93e-3(5.29e-4) &	5.90e-3(5.27e-4) &	7.40e-3(5.31e-4) \\
    \bottomrule[1pt]
\end{tabularx}
% \vspace{-2.25em}
\end{table}

\begin{table}[h]
% \vspace{-1.5em}
\caption{Scalability - different number of contacts}
\label{tab:scale-contacts}
\centering
\small
\begin{tabularx}{0.90\textwidth}{c | Y | Y | Y | Y }
    \toprule[1pt]
     & CM-CD-CLNN & MLP-CD-CLNN & IN-CP-CLNN & IN-SP-CP  \\
     \midrule[0.5pt]
     D=100 & 1.97e-3(4.72e-4) &	1.85e-2(2.53e-3) &	2.31e-2(2.12e-3) &	2.93e-2(2.05e-3) \\
     D=200 & 9.39e-4(3.26e-4) &	1.08e-2(8.91e-4) &	1.27e-2(7.76e-4) &	1.37e-2(8.14e-4) \\
     D=400 & 3.20e-3(5.69e-4) &	5.98e-3(5.21e-4) &	6.87e-3(5.11e-4) &	8.94e-3(5.30e-4) \\
    \bottomrule[1pt]
\end{tabularx}
% \vspace{-2.25em}
\end{table}

\end{document}